\documentclass{article}

\usepackage{microtype}
\usepackage{graphicx}
\usepackage{subcaption}
\usepackage{booktabs} 
\usepackage[table]{xcolor} 
\usepackage{colortbl} 

\usepackage{hyperref}
\usepackage{enumitem}

\usepackage[preprint]{icml2026}

\usepackage{amsmath}
\usepackage{amssymb}
\usepackage{mathtools}
\usepackage{amsthm}

\definecolor{bestperf}{RGB}{220,235,255}  
\definecolor{poorperf}{RGB}{255,230,230}  

\definecolor{mhaBlue}{RGB}{68, 114, 196}
\definecolor{mhaFill}{RGB}{218, 227, 243}
\definecolor{sharedGreen}{RGB}{0, 176, 80}
\definecolor{sharedFill}{RGB}{226, 240, 217}
\definecolor{residOrange}{RGB}{237, 125, 49}
\definecolor{residFill}{RGB}{252, 228, 214}

\usepackage[capitalize,noabbrev]{cleveref}

\theoremstyle{plain}

\theoremstyle{definition}

\theoremstyle{remark}

\usepackage[textsize=tiny]{todonotes}

\icmltitlerunning{Low-Rank Key Value Attention}

\begin{document}

\twocolumn[
  \icmltitle{Low-Rank Key Value Attention}



  \icmlsetsymbol{equal}{*}

\begin{icmlauthorlist}
  \icmlauthor{James O'Neill$^\dagger$}{intercom}
  \icmlauthor{Robert Clancy}{intercom}
  \icmlauthor{Mariia Matskevichus}{intercom}
  \icmlauthor{Fergal Reid}{intercom}
\end{icmlauthorlist}

\icmlaffiliation{intercom}{
  AI Group, Intercom\\
  124 St Stephen's Green, Dublin 2, D02 C628, Ireland
}

\icmlcorrespondingauthor{James O'Neill}{james.oneill@intercom.io}

  \icmlkeywords{Machine Learning, ICML}

  \vskip 0.3in
]



\printAffiliationsAndNotice{}  

\begin{abstract}
The key-value (KV) cache is a primary memory bottleneck in Transformers. We propose Low-Rank Key-Value (LRKV) attention, which reduces KV cache memory by exploiting redundancy across attention heads, while being compute efficient. Each layer uses a shared full-rank KV projection augmented with low-rank, head-specific residuals, providing a continuous trade-off between complete sharing and full independence.                                                                                                  
After pretraining models of size 128M to 6.3B parameters, LRKV consistently achieves the lowest test loss among standard MHA, MQA/GQA, and MLA while using only 45-53\% of MHA's KV cache. LRKV reaches equivalent baseline quality 18-25\% faster (measured in training steps). After supervised midtraining, LRKV achieves the highest downstream task performance across ARC-Easy, ARC-Challenge, MMLU, GSM8K, and HumanEval benchmarks. 
\end{abstract}

\section{Introduction}
Transformers are the dominant architecture for large-scale sequence modeling in language, vision, and multimodal domains~\citep{vaswani2017attention,openai2023gpt4}, but as their size, sequence length, and context window grow, so does, rapidly, their computational and memory costs.  KV-caching, which stores the attention key and value representations, is a primary contributor to this overhead, as it spans every attention layer, and scales linearly with sequence length and count of heads. The cumulative KV footprint of modern models with tens of billion parameters can exceed the parameter memory itself, especially for long-context inference~\citep{dao2024flashattention3,child2019generating}.
To illustrate the scale of this challenge, consider a 2.5B parameter model (18 layers, 18 heads, $d_h{=}128$) serving 8K-token contexts: the KV cache alone requires $2 \times 18 \times 18     \times 128 \times 8192 \times 2\text{ bytes} = 648\text{ MB}$ per request. At batch size 4, KV cache memory (2.6 GB) exceeds the model's parameter memory (5 GB in bfloat16), consuming more   
memory than the model itself. This forces practitioners into an impossible trilemma: use smaller batches (sacrificing throughput), shorter contexts (limiting capability), or frequent cache eviction (adding latency)—each option directly degrading production utility. Reducing KV cache size thus directly translates to 2$\times$ longer contexts or 2$\times$ larger batches at fixed memory budgets.
\begin{figure*}[ht]
    \centering
    \includegraphics[width=.75\linewidth]{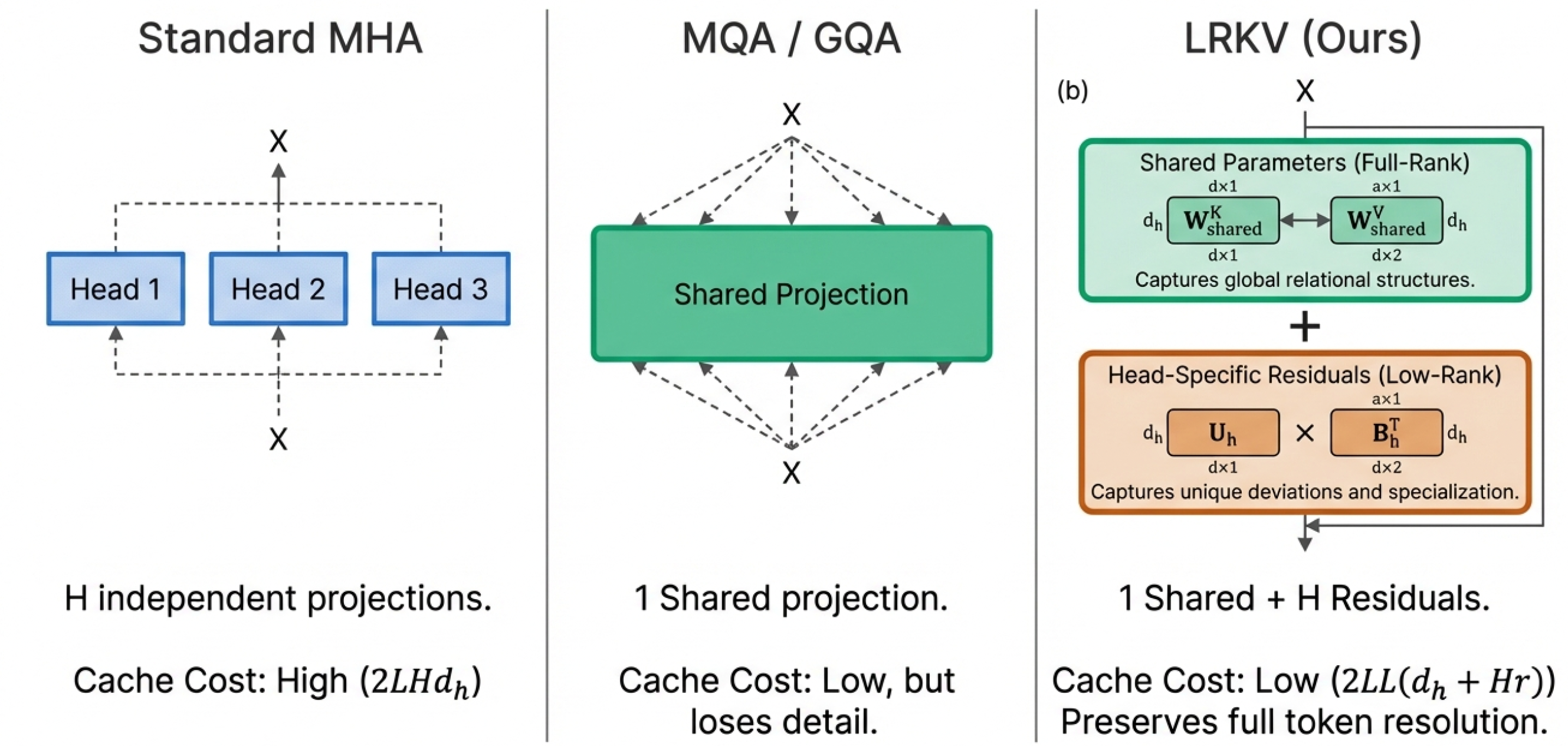}
\caption{Comparison of attention mechanisms. \textbf{Standard MHA:} Uses $H$ independent projections per head, requiring high cache cost ($2LHd_h$). \textbf{MQA/GQA:} Shares projections across all heads, reducing cache but losing head-specific detail. \textbf{LRKV (Ours):} Combines \textcolor{sharedGreen}{shared full-rank parameters} with \textcolor{residOrange}{head-specific low-rank residuals} ($\mathbf{U}_h \mathbf{B}_h^T$, rank $r \ll d_h$), achieving low cache cost $2L(d_h + Hr)$ while preserving head diversity. 
}
\label{fig:attention_comparison}
\end{figure*}


A range of approaches have been proposed to alleviate the growing KV cache cost. Multi-Query Attention~\citep[MQA]{shazeer2019fast} and Grouped-Query Attention~\citep[GQA]{ainslie2023gqa} reduce memory and latency by sharing key-value (KV) projections across heads or groups of heads, and are now standard in large-scale models such as PaLM and LLaMA~\citep{touvron2023llama2,touvron2024llama3}.
However, aggressive KV sharing introduces a fundamental trade-off: while memory is reduced, head-level representational diversity is constrained, even though distinct attention heads are known to encode complementary syntactic and semantic patterns~\citep{clark2019does,michel2019sixteen}.

At the same time, empirical studies and recent spectral analyses show that attention heads are not fully independent: head-specific KV projections are highly correlated and occupy overlapping subspaces, indicating substantial redundancy~\citep{yunis2024approaching}.
Crucially, this redundancy is structured rather than uniform—small, head-specific variations remain important for capturing nuanced dependencies.
This raises a natural question: \emph{can KV memory be reduced by exploiting redundancy across heads, without collapsing the specialization that makes multi-head attention effective?}

Beyond KV sharing, efficiency research has pursued complementary directions such as sparse or kernelized attention~\citep{wang2020linformer,beltagy2020longformer,child2019generating}, architectural optimizations~\citep{dao2024flashattention3}, and latent compression methods including Multi-Latent Attention~\citep[MLA]{dao2024mlattention}. However, none of these approaches explicitly resolve the duplication of per-head key and value representations that dominates the KV memory footprint in large models.

In this work, we propose \textbf{Low-Rank Key-Value Attention (LRKV)}, a simple modification of multi-head attention that exploits structured redundancy across heads while preserving head specialization. Each Transformer layer maintains a shared, full-rank key and value projection that serves as a global basis, while each head learns a compact, trainable low-rank residual that captures head-specific deviations. The shared component encodes global relational structure, and the residuals restore the localized diversity otherwise lost under aggressive KV sharing. This factorization substantially reduces KV memory while retaining the expressivity of multi-head attention.

LRKV differs fundamentally from prior KV-efficient designs in where compression is applied. Unlike LoRA~\citep{hu2022lora}, which introduces low-rank structure post hoc during fine-tuning, LRKV learns head-specific low-rank residuals jointly during pretraining. Unlike factorized QKV methods that reduce rank along the token or hidden dimension~\citep{wang2020linformer,saxena2024eigen,chang2025palu},
LRKV preserves full token resolution and compresses only across attention heads. We now discuss the main contributions of this paper.

\textbf{Contributions.}(1)~LRKV reduces KV cache by 45-53\% while achieving lowest test loss across 128M-6.3B models, reaching baseline performance 18-25\% faster in training steps.
(2)~LRKV achieves highest downstream performance across ARC, MMLU, GSM8K, demonstrating better pretraining translates to improved capabilities.
(3)~Gauge-invariant analysis shows LRKV effectively uses low-rank structure by preserving 93.5\% head diversity vs 94\% MHA.\footnote{Code and trained models will be released upon acceptance.}

\section{Related Work}

\textbf{KV-Sharing Mechanisms.} MQA and GQA reduce KV cache by sharing projections across heads, but sacrifice head-level expressivity. MQA~\citep{shazeer2019fast} uses a single shared KV projection for all heads, achieving maximal cache reduction but constraining diversity. GQA~\citep{ainslie2023gqa} groups heads to share KV projections within groups, interpolating between MQA and full MHA. While both methods reduce memory, they fundamentally limit the representational capacity available to each head. LRKV preserves shared projection efficiency while restoring diversity through structured low-rank head residuals, achieving better quality at comparable cache sizes.

\textbf{Latent Compression Approaches.} MLA~\citep{dao2024mlattention} takes a complementary approach: rather than sharing across heads, it compresses across the token dimension by projecting inputs into a low-dimensional latent space ($d_c \ll d$) before caching. While effective for extreme compression, this bottleneck constrains all heads to operate through the same latent representation and requires T-dependent reconstruction overhead during generation (see ~\autoref{appendix:hyperparams}). LRKV instead preserves full token-level resolution and compresses along the head dimension via additive factorization ($\mathbf{W}_h = \mathbf{W}_{\text{shared}} + \mathbf{U}_h\mathbf{B}_h^\top$), avoiding latent bottlenecks while maintaining per-head specialization. These represent orthogonal design choices: MLA optimizes for minimal cache size via aggressive token compression, while LRKV balances memory efficiency with head-level expressivity.

\textbf{Low-Rank Parameterization.} LoRA~\citep{hu2022lora} introduces low-rank updates for parameter-efficient fine-tuning, intervening on the optimization pathway rather than the cached representations. LRKV applies the low-rank principle directly to KV projections during pretraining, shaping the structure of cached features from the start. Recent work explores low-rank QKV factorizations~\citep{xie2023gptq,khalaf2025qkv,lv2024scalable} to reduce parameters or computation; LRKV differs by preserving a full-rank shared base (capturing global structure) while using low-rank residuals only for head-specific deviations. Concurrent work on factorized KV mechanisms (MFA/MFA-KR~\citep{hu2025mfa}, TPA~\citep{zhang2025tpa}) explores related structured compression ideas.

Unlike joint-head low-rank decompositions such as J-LRD/Palu, which compress projections post hoc, LRKV is a pretraining-time architectural reparameterization that preserves a full-rank shared base while learning head-specific low-rank residuals. This enables exact associative decoding and reduces KV cache without post-hoc approximation.

\textbf{Complementary Efficiency Methods.} LRKV is orthogonal to other efficiency techniques: sparse/kernelized attention~\citep{child2019generating,beltagy2020longformer,wang2020linformer} reduce computational complexity but not KV cache size; architectural optimizations like FlashAttention~\citep{dao2024flashattention3} improve throughput through better memory access patterns; quantization and pruning reduce precision or parameters. For head diversity analysis, we extend prior metrics~\citep{michel2019sixteen,kornblith2019similarity,wang2025complete} using gauge-invariant bilinear forms with centered Gram matrix analysis (see~\autoref{appendix:related_work}).

\section{Methodology}\label{sec:methodology}
Transformers represent each token in a sequence through queries, keys, and values that interact via scaled dot-product attention. For an input matrix $\mathbf{X} \in \mathbb{R}^{L \times d}$, the $h$-th attention head computes $\mathbf{Q}_h = \mathbf{X} \mathbf{W}_h^Q$, $\mathbf{K}_h = \mathbf{X} \mathbf{W}_h^K$, and $\mathbf{V}_h = \mathbf{X} \mathbf{W}_h^V$, where $\mathbf{W}_h^{Q,K,V} \in \mathbb{R}^{d \times d_h}$ and $d_h = d/H$ for $H$ heads. The head output is
\begin{equation}
\mathbf{O}_h
=
\mathrm{softmax}\!\left(\frac{\mathbf{Q}_h \mathbf{K}_h^\top}{\sqrt{d_h}}\right)\mathbf{V}_h,
\label{eq:standard_attention}
\end{equation}
and outputs from all heads are concatenated and projected to form the layer output.

\textbf{KV caching.}
During autoregressive decoding, the KV cache stores $\mathbf{K}_h$ and $\mathbf{V}_h$ for all previous tokens and heads, incurring per-layer memory
$M_{\text{standard}} = 2 L H d_h = 2Ld$.
Equivalently, per token the cache stores $2Hd_h$ floating-point values (keys and values across $H$ heads). Over $N$ layers in bfloat16 precision, this is $2N \cdot (2LHd_h)$ bytes of KV storage.

\textbf{Motivation.}
Empirical studies show attention heads within a layer are often correlated ~\citep{clark2019does,michel2019sixteen}, suggesting that per-head key/value features contain substantial redundancy. Existing KV-sharing methods such as MQA and GQA reduce cache size by sharing K/V across heads, but can reduce head diversity and modeling capacity. Our goal is to reduce redundant KV storage while preserving head-specific flexibility.

\textbf{Low-Rank KV Attention.}
We parameterize each head's key/value projection as a shared dense (unconstrained) base plus a head-specific low-rank residual, while keeping the resulting projection shape $d \times d_h$:
\begin{align}
\mathbf{W}_h^K &= \mathbf{W}_{\mathrm{shared}}^K + \mathbf{U}_h^K {\mathbf{B}_h^K}^{\top}, \label{eq:lrkv_weights_k} \\
\mathbf{W}_h^V &= \mathbf{W}_{\mathrm{shared}}^V + \mathbf{U}_h^V {\mathbf{B}_h^V}^{\top},
\label{eq:lrkv_weights}
\end{align}
where $\mathbf{W}_{\mathrm{shared}}^{K,V}\in\mathbb{R}^{d\times d_h}$ are dense projection matrices with no rank constraint, $\mathbf{U}_h^{K,V}\in\mathbb{R}^{d\times r}$, $\mathbf{B}_h^{K,V}\in\mathbb{R}^{d_h\times r}$, and $r\ll d_h$.
The effective keys and values are $\mathbf{K}_h = \mathbf{X}\mathbf{W}_h^K$ and $\mathbf{V}_h = \mathbf{X}\mathbf{W}_h^V$.
When $r=0$, LRKV reduces to complete KV sharing (MQA-style) within a layer; increasing $r$ interpolates continuously toward standard MHA.

\textbf{Training and initialization.}
All parameters in \autoref{eq:lrkv_weights} are optimized jointly during pretraining. Gradients from the shared and residual paths are additive, allowing the model to learn how much per-head variation is required.
We initialize $\mathbf{W}_{\mathrm{shared}}^{K,V}$ using standard Kaiming initialization for attention projections. The per-head factors $\mathbf{U}_h^{K,V}$ and $\mathbf{B}_h^{K,V}$ are initialized to small random values scaled by $1/\sqrt{r}$ such that initial residuals $\mathbf{U}_h {\mathbf{B}_h}^\top$ have magnitude $\approx 0.1 \times \|\mathbf{W}_{\mathrm{shared}}\|_F$, ensuring the model starts close to the shared baseline and gradually learns head specialization.
During training, we materialize full $\mathbf{K}_h$ and $\mathbf{V}_h$ matrices before attention computation for simplicity; the factored form is used only during inference to realize memory savings. This is an exact reparameterization, so the factorized and unfactorized forms are mathematically equivalent.
We apply low-rank factorization only to key and value projections, as these are cached during inference; query projections remain independent per head at full rank.

\textbf{LRKV caching scheme.}
During decoding, the bottleneck is storing $\mathbf{K}_h,\mathbf{V}_h\in\mathbb{R}^{L\times d_h}$ for every head. LRKV instead caches shared features once per layer,
\[
\mathbf{K}_{\mathrm{shared}} = \mathbf{X} \mathbf{W}_{\mathrm{shared}}^K,\quad
\mathbf{V}_{\mathrm{shared}} = \mathbf{X} \mathbf{W}_{\mathrm{shared}}^V \in\mathbb{R}^{L\times d_h},
\]
and compact per-head latents
\[
\mathbf{R}_h^K = \mathbf{X} \mathbf{U}_h^K,\quad
\mathbf{R}_h^V = \mathbf{X} \mathbf{U}_h^V \in\mathbb{R}^{L\times r}.
\]
The full per-head features implied by \autoref{eq:lrkv_weights} are
\begin{align}
\mathbf{K}_h &= \mathbf{K}_{\mathrm{shared}} + \mathbf{R}_h^K {\mathbf{B}_h^K}^{\top}, \label{eq:lrkv_reconstruct_k} \\
\mathbf{V}_h &= \mathbf{V}_{\mathrm{shared}} + \mathbf{R}_h^V {\mathbf{B}_h^V}^{\top}.
\label{eq:lrkv_reconstruct}
\end{align}
Importantly, the matrices $\mathbf{B}_h^{K,V}$ are model parameters and are \emph{not} cached per token.

\textbf{Attention computation without explicit reconstruction.}
Naively materializing $\mathbf{K}_h,\mathbf{V}_h$ for all cached tokens would cost $O(Lrd_h)$ memory and compute. Instead, we exploit associativity to compute attention exactly without forming full per-head KV tensors. For a decoding step with query $\mathbf{q}_h\in\mathbb{R}^{d_h}$,
\begin{equation}
\mathbf{q}_h \mathbf{K}_h^\top
=
\mathbf{q}_h \mathbf{K}_{\mathrm{shared}}^\top 
+ (\mathbf{q}_h \mathbf{B}_h^K)\,(\mathbf{R}_h^K)^\top,
\label{eq:lrkv_logits}
\end{equation}
and for attention weights $\mathbf{a}_h\in\mathbb{R}^{1\times L}$,
\begin{equation}
\mathbf{a}_h \mathbf{V}_h
=
\mathbf{a}_h \mathbf{V}_{\mathrm{shared}} 
+ (\mathbf{a}_h \mathbf{R}_h^V)\,{\mathbf{B}_h^V}^{\top}.
\label{eq:lrkv_values}
\end{equation}
Equations~\ref{eq:lrkv_logits} and \ref{eq:lrkv_values} compute the \emph{exact} attention logits and outputs implied by \autoref{eq:lrkv_reconstruct_k}--\autoref{eq:lrkv_reconstruct}, but avoid explicit reconstruction of $\mathbf{K}_h,\mathbf{V}_h$. This form can be implemented inside fused attention kernels. See~\autoref{tab:attention_math_comparison} for a detailed mathematical comparison to existing mechanisms.

\paragraph{Compatibility with Positional Embeddings.}
Positional embeddings, such as RoPE, are applied after projection and distribute linearly over the LRKV decomposition:
\[
\mathrm{RoPE}(\mathbf{K}_h) = \mathrm{RoPE}(\mathbf{K}_{\mathrm{shared}}) + \mathrm{RoPE}(\mathbf{R}_h)\mathbf{B}_h^\top.
\]
Both components are rotated once at caching time, as in standard KV caching, and reused during decoding. This introduces no additional memory or sequence-length-dependent computation. In contrast to latent-attention approaches such as MLA, which require partial RoPE to preserve projection absorption, LRKV supports full-dimension RoPE without modification.

\textbf{KV cache memory complexity (during decoding).}
Standard attention stores per-head keys and values:
\[
M_{\text{standard}} = 2 L H d_h.
\]
LRKV stores shared features plus per-head latents:
\begin{align}
M_{\text{LRKV}}
&= \underbrace{2Ld_h}_{\text{shared }(K,V)}
+
\underbrace{2LHr}_{\text{per-head latents}} \nonumber \\
&= 2L(d_h + Hr).
\label{eq:lrkv_memory}
\end{align}
Thus the memory ratio is
\begin{equation}
\frac{M_{\text{LRKV}}}{M_{\text{standard}}}
=
\frac{d_h + Hr}{H d_h}
=
\frac{1}{H} + \frac{r}{d_h}.
\label{eq:lrkv_ratio}
\end{equation}
Per token, LRKV stores $2d_h + 2Hr$ values versus $2Hd_h$ for standard attention.
\begin{table*}[t]
\centering
\caption{\textbf{Pretraining performance across model scales.} LRKV achieves competitive test loss across all model scales (128M, 1.2B, 2.5B, 6.3B) while maintaining efficient KV cache usage.}
\label{tab:pretraining_results}
\resizebox{\textwidth}{!}{%
\small
\begin{tabular}{l|c|cccc|cc|cc|cc|cc}
\toprule
& \textbf{Arch.} & \multicolumn{4}{c|}{\textbf{KV Cache \%}} & \multicolumn{2}{c|}{\textbf{128M}} & \multicolumn{2}{c|}{\textbf{1.2B}} & \multicolumn{2}{c|}{\textbf{2.5B}} & \multicolumn{2}{c}{\textbf{6.3B}} \\
\cmidrule(lr){2-2} \cmidrule(lr){3-6} \cmidrule(lr){7-8} \cmidrule(lr){9-10} \cmidrule(lr){11-12} \cmidrule(lr){13-14}
\textbf{Model} & \textbf{KV Heads} & \textbf{128M} & \textbf{1.2B} & \textbf{2.5B} & \textbf{6.3B} & \textbf{CE} $\downarrow$ & \textbf{BPB} $\downarrow$ & \textbf{CE} $\downarrow$ & \textbf{BPB} $\downarrow$ & \textbf{CE} $\downarrow$ & \textbf{BPB} $\downarrow$ & \textbf{CE} $\downarrow$ & \textbf{BPB} $\downarrow$ \\
\midrule
Standard MHA & 6/12/18/32 & 100 & 100 & 100 & 100 & 2.903 & 0.878 & 2.530 & 0.765 & 2.389 & 0.723 & 2.319 & 0.701 \\
GQA & 3/4/6/2 & 50 & 33 & 33 & 6 & 2.918 & 0.883 & 2.536 & 0.767 & 2.397 & 0.725 & 2.354 & 0.712 \\
MQA & 1 & 17 & 8 & 6 & 3 & 2.929 & 0.886 & 2.573 & 0.778 & 2.408 & 0.729 & 2.304 & 0.697 \\
MLA & -- & 8.3 & 8.3 & 8.3 & 12.5 & 2.901 & 0.876 & 2.564 & 0.775 & 2.392 & 0.724 & 2.377 & 0.719 \\
\textbf{LRKV} & -- & 53 & 48 & 48 & 45 & \cellcolor{bestperf}\underline{\textbf{2.893}} & \cellcolor{bestperf}\textbf{0.872} & \cellcolor{bestperf}\underline{\textbf{2.509}} & \cellcolor{bestperf}\textbf{0.758} & \cellcolor{bestperf}\underline{\textbf{2.376}} & \cellcolor{bestperf}\textbf{0.719} & \cellcolor{bestperf}\underline{\textbf{2.288}} & \cellcolor{bestperf}\textbf{0.692} \\
\bottomrule
\end{tabular}%
}
\caption*{\small KV Heads show values for 128M/1.2B/2.5B/6.3B scales. KV Cache \% indicates memory relative to Standard MHA baseline (see~\autoref{appendix:memory_profiling} for profiling details). Test metrics on held-out FineWeb-Edu (100B tokens for 128M, 50B for 1.2B/2.5B/6.3B).}
\end{table*}

\textbf{Compute overhead during decoding (FLOPs).}
Per decoding step, standard attention has dominant per-head computational cost $O(Ld_h)$ from query--key dot products and value aggregation. Using the distributed forms in \autoref{eq:lrkv_logits} and \autoref{eq:lrkv_values}, LRKV adds an extra $\Delta \mathrm{FLOPs} = O(Lr + rd_h)$ per head per decoding step. For long contexts ($L\gg 1$), the dominant additional term scales as $O(Lr)$, yielding a relative overhead of approximately $r/d_h$ compared to standard attention.

\textbf{Memory bandwidth.}
Modern inference is often bandwidth-bound rather than compute-bound ~\citep{dao2024flashattention3}. Per decoding step, standard MHA reads $2H$ cached tensors ($\mathbf{K}_h$ and $\mathbf{V}_h$ for each head). LRKV reads two shared tensors plus $2H$ per-head latent tensors. Since the shared tensors ($L \times d_h$) are reused across heads and the per-head latents ($L \times r$) are substantially smaller ($r \ll d_h$), the total bytes transferred is $2L(d_h + Hr)$ for LRKV versus $2LHd_h$ for standard MHA, matching the cache-size reduction and directly translating memory savings into reduced bandwidth pressure.

\textbf{Parameter complexity.}
Standard per-layer K/V parameters scale as $P_{\text{MHA,KV}} = 2H d d_h$.
LRKV uses a shared base plus low-rank factors:
\begin{equation}
P_{\text{LRKV}}
=
\underbrace{2 d d_h}_{\text{shared }(K,V)}
+
\underbrace{2H r(d + d_h)}_{\text{per-head low-rank factors}},
\label{eq:lrkv_params}
\end{equation}
which may be lower or higher than standard depending on $(H,r,d_h)$. In practice, we choose $r$ to prioritize KV-cache reduction with minimal quality impact; in our experiments, LRKV remains competitive or superior even when the total K/V parameter count is below that of standard MHA.

\textbf{Rank as a control for diversity.}
The residual rank $r$ controls a continuous spectrum between fully shared KVs (MQA-style, $r=0$) and fully independent per-head projections (standard MHA, large $r$). The shared base $\mathbf{W}_{\mathrm{shared}}^{K,V}$ captures globally useful features, while low-rank residuals enable head specialization within a constrained budget. Empirically, we find $r \approx 0.36$-$0.43 \times d_h$ provides a critical threshold: LRKV achieves 93.5\% PCA-based effective rank versus 94.0\% for standard MHA at 2.5B scale (\autoref{sec:theory_results}), preserving nearly all head diversity while achieving substantial cache reduction. See~\autoref{appendix:spectral_structure} for detailed discussion of spectral structure and low-rank factorization.

\begin{figure*}[h]
\centering
\includegraphics[width=0.95\linewidth]{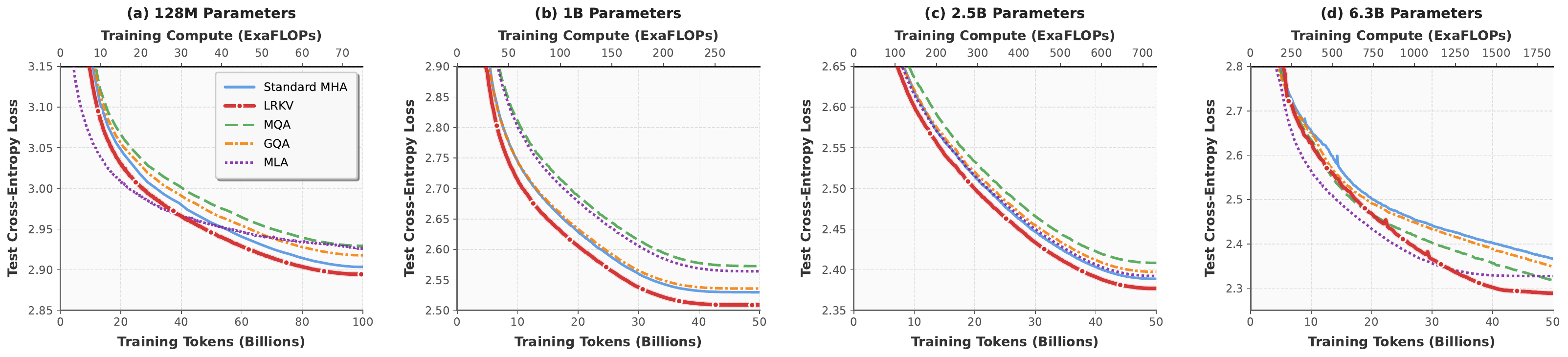}
\caption{\textbf{Cross-scale pretraining curves with dual-axis compute metrics.}
Test cross-entropy loss across four model scales (128M, 1.2B, 2.5B, 6.3B) plotted against training tokens (bottom x-axis) and cumulative training compute in ExaFLOPs (top x-axis). LRKV demonstrates competitive performance across all scales, achieving the lowest test loss at 128M and 2.5B scales. Final results in~\autoref{tab:pretraining_results}.}
\label{fig:loss_curves}
\end{figure*}
\section{Experiments}\label{sec:results}

We evaluate LRKV through large-scale pretraining experiments across four model sizes (128M, 1.2B, 2.5B, 6.3B parameters), comparing against standard MHA and state-of-the-art KV-efficient baselines. We measure pretraining loss, training efficiency, downstream task performance, and provide detailed analysis of why LRKV preserves modeling quality despite substantial cache reduction.

\textbf{Experimental setup.}
We pretrain decoder-only Transformers at four scales (128M, 1.2B, 2.5B, 6.3B) on FineWeb-Edu~\citep{fineweb2024} for 100B tokens (128M) and 50B tokens (others), comparing LRKV against Standard MHA, GQA, MQA, and MLA. Models use 2048-token context, Muon+AdamW optimizers~\citep{muon2024}, and are pretrained on 8$\times$H200 GPUs in bfloat16. After pretraining, we perform supervised midtraining on 568K examples from SmolTalk~\citep{allal2024smoltalk}, MMLU~\citep{hendrycks2021mmlu}, and GSM8K~\citep{cobbe2021gsm8k}, then evaluate on ARC-Easy, ARC-Challenge, MMLU, GSM8K, and HumanEval. LRKV uses 45-53\% of Standard MHA's KV cache (see~\autoref{appendix:experimental_setup} for details). 

\subsection{Pretraining Results}
\paragraph{Final pretraining performance.}
~\autoref{tab:pretraining_results} shows LRKV achieves competitive test loss across all scales (128M, 1.2B, 2.5B, 6.3B), with the best performance at 128M and 2.5B scales. At 6.3B scale, LRKV reaches 2.288 CE (0.692 BPB), outperforming MHA (2.319 CE), MQA (2.304 CE), GQA (2.354 CE), and MLA (2.377 CE). At 1.2B scale, LRKV achieves 2.509 CE (0.758 BPB) with only 48\% of MHA's cache—outperforming all baselines while using half the memory. At 2.5B, LRKV achieves 0.719 BPB with only 48.4\% of MHA's cache—a strictly better accuracy-memory tradeoff. Cache efficiency improves at larger scales (52.6\% → 48\% → 48.4\% → 45.1\%), making LRKV increasingly attractive for large models.
\begin{table*}[t]
\centering
\caption{\textbf{Downstream task performance after midtraining (128M, 2.5B, and 6.3B scales).} LRKV achieves the highest combined accuracy across all three scales (18.9\%, 37.9\%, 40.2\%) on five diverse benchmarks, demonstrating that superior pretraining performance translates to stronger downstream capabilities. \small Combined score (Comb.) is the average across all five benchmarks. HE = HumanEval. }
\label{tab:midtraining_results}
\resizebox{\textwidth}{!}{%
\small
\begin{tabular}{l|ccccc|c|ccccc|c|ccccc|c}
\toprule
& \multicolumn{6}{c|}{\textbf{128M}} & \multicolumn{6}{c|}{\textbf{2.5B}} & \multicolumn{6}{c}{\textbf{6.3B}} \\
\cmidrule(lr){2-7} \cmidrule(lr){8-13} \cmidrule(lr){14-19}
\textbf{Model} & \textbf{ARC-E} & \textbf{ARC-C} & \textbf{MMLU} & \textbf{GSM8K} & \textbf{HE} & \textbf{Comb.} & \textbf{ARC-E} & \textbf{ARC-C} & \textbf{MMLU} & \textbf{GSM8K} & \textbf{HE} & \textbf{Comb.} & \textbf{ARC-E} & \textbf{ARC-C} & \textbf{MMLU} & \textbf{GSM8K} & \textbf{HE} & \textbf{Comb.} \\
\midrule
Standard MHA & 26.4 & 26.0 & 27.2 & 1.0 & 8.8 & 17.9 & 66.6 & 47.1 & 39.3 & 10.2 & \textbf{13.4} & 35.3 & 72.7 & 53.6 & 42.7 & 10.6 & 14.6 & 38.8 \\
GQA & 28.0 & 25.4 & 25.7 & 0.5 & 4.6 & 16.8 & 65.4 & 49.6 & 40.5 & 10.0 & \textbf{13.4} & 35.8 & 69.0 & 48.2 & 40.9 & 9.9 & 13.4 & 36.3 \\
MQA & 28.0 & 27.1 & 27.8 & 1.3 & \cellcolor{poorperf}2.4 & 17.3 & 65.2 & 47.3 & 40.2 & 10.3 & \cellcolor{poorperf}3.7 & 33.3 & 67.2 & 47.0 & 40.5 & 8.2 & \cellcolor{poorperf}4.3 & 33.4 \\
MLA & 27.3 & 27.8 & 26.3 & 1.1 & 7.7 & 18.0 & 67.5 & 51.5 & 41.9 & 10.8 & 12.8 & 36.9 & 69.4 & 51.3 & 40.9 & 9.5 & 13.4 & 36.9 \\
\textbf{LRKV} & 26.7 & \cellcolor{bestperf}\textbf{30.2} & \cellcolor{bestperf}\textbf{28.4} & 0.3 & \cellcolor{bestperf}\textbf{9.1} & \cellcolor{bestperf}\underline{\textbf{18.9}} & \cellcolor{bestperf}\textbf{70.7} & \cellcolor{bestperf}\textbf{53.8} & \cellcolor{bestperf}\textbf{42.2} & \cellcolor{bestperf}\textbf{11.3} & 11.7 & \cellcolor{bestperf}\underline{\textbf{37.9}} & \cellcolor{bestperf}\textbf{75.0} & \cellcolor{bestperf}\textbf{58.0} & \cellcolor{bestperf}\textbf{44.5} & \cellcolor{bestperf}\textbf{11.5} & 12.8 & \cellcolor{bestperf}\underline{\textbf{40.2}} \\
\bottomrule
\end{tabular}%
}
\end{table*}


\textbf{Cross-scale training efficiency and compute normalization.}
~\autoref{fig:loss_curves} presents test performance across all four model scales, measured in cross-entropy loss with dual-axis compute metrics (training tokens and cumulative FLOPs).

LRKV demonstrates competitive performance across scales, achieving the lowest test loss at 128M and 2.5B scales while remaining highly competitive at 1.2B and 6.3B, showing consistent advantages across three orders of magnitude in model size. The top x-axis shows cumulative training compute (ExaFLOPs), accounting for mechanism-specific per-token costs (MQA -6\%, GQA -3.2\%, MLA -3.7\%, LRKV +0.8\% at 2.5B scale).\footnote{The reported per-token costs reflect \emph{inference} FLOPs. During training, LRKV materializes full K/V matrices (as do all methods), so training cost per step is comparable to MHA. The 18-25\% ``training compute savings'' refer to sample efficiency (fewer steps to reach target loss), not per-step speedup.} When normalized by total compute rather than token count, LRKV maintains its performance advantage: at every ExaFLOP milestone, LRKV achieves lower test loss than all baselines. At 2.5B scale, LRKV reaches equivalent baseline performance 18--30\% faster (23.6\% average across all methods), while no baseline reaches LRKV's final performance even after exhausting the full compute budget. This establishes LRKV as dominant in both sample efficiency and final quality. At the largest 6.3B scale, the performance gap widens: LRKV achieves 2.288 CE versus Standard MHA's 2.319 CE (1.3\% improvement), demonstrating that LRKV's architectural advantages strengthen rather than diminish with scale.
\begin{figure}[h]
\centering
\includegraphics[width=1.0\linewidth]{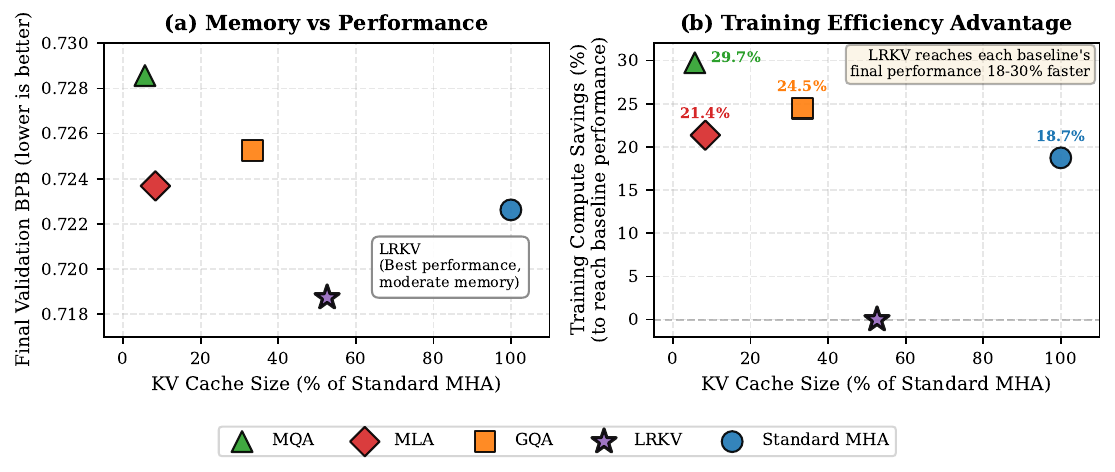}
\caption{\textbf{LRKV achieves superior training efficiency alongside best performance (2.5B scale).}
\small{\textbf{(a) Memory vs Performance:}  Test BPB versus KV cache percentage for all methods. LRKV achieves optimal trade-off with lowest BPB at 48.4\% cache usage (2.5B scale).
\textbf{(b) Training Efficiency Advantage:} LRKV reaches each baseline's final test loss, quantifying training compute savings. LRKV reaches all baselines' performance earlier.}}
\label{fig:training_efficiency}
\end{figure}

\textbf{Training efficiency analysis.}
Beyond superior converged performance, LRKV demonstrates remarkable sample efficiency at 2.5B scale (\autoref{fig:training_efficiency}). LRKV reaches each baseline's final validation performance 18-30\% faster, averaging \textbf{23.6\% training compute savings} across all baselines while achieving better final performance. Critically, this reveals an asymmetric advantage: LRKV reaches any baseline's performance target early in training, but no baseline reaches LRKV's final performance (0.719 BPB) even after the full 50B token budget.

\begin{figure}[t]
\centering
\includegraphics[width=1.0\linewidth]{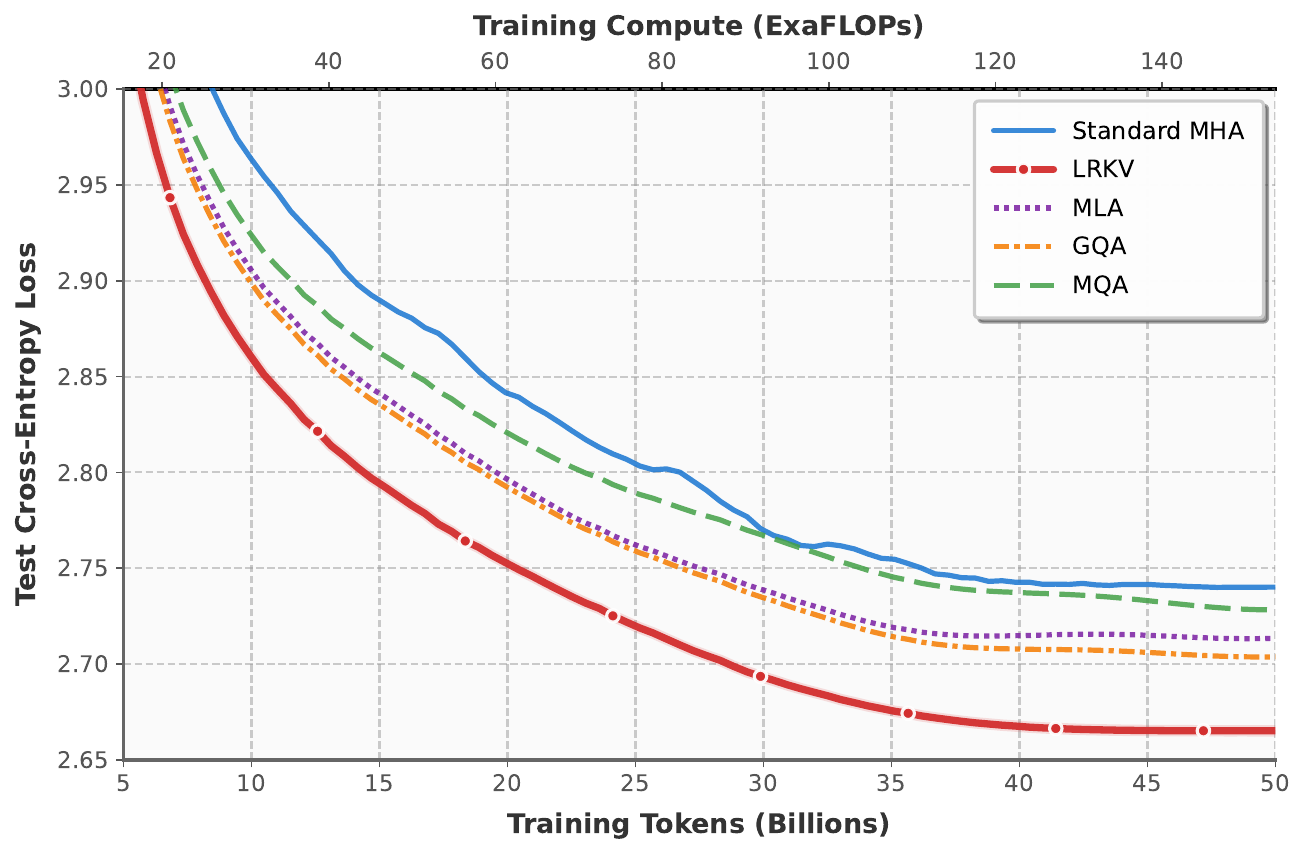}
\caption{\textbf{Long context pretraining for 512M parameter models.} Test cross-entropy loss curves for models trained with 8192-token sequence length on 50B tokens (153.6 ExaFLOPs). LRKV outperforms Standard MHA by 2.7\% while using 48.2\% cache.}\label{fig:512m_8k_context}
\end{figure}

\textbf{Long Context Pretraining.} We extend evaluation to 8192 token sequences using 512M parameter models trained for 50B tokens (153.6 ExaFLOPs). \autoref{fig:512m_8k_context} shows that LRKV achieves 2.67 test loss, outperforming MHA (2.74) by 2.7\% while using only 48.2\% of its KV cache. LRKV also surpasses GQA (2.70, -1.3\%), MLA (2.71, -0.9\%), and MQA (2.73, -0.4\%). Notably, all KV-efficient methods outperform Standard MHA at long context, suggesting compression provides implicit regularization benefits, though LRKV's architectural advantages remain pronounced. These results validate LRKV's effectiveness at extended sequence lengths.

\subsection{Downstream Task Performance}
To evaluate whether LRKV's pretraining advantages translate to practical capabilities, we perform supervised midtraining on a diverse instruction-following dataset (568K examples from SmolTalk~\citep{allal2024smoltalk}, MMLU~\citep{hendrycks2021mmlu}, and GSM8K~\citep{cobbe2021gsm8k}) and evaluate on five standard benchmarks.~\autoref{tab:midtraining_results} shows final downstream performance across three scales (128M, 2.5B, 6.3B).
\begin{figure*}[ht]
\centering
\includegraphics[width=1.0\linewidth]{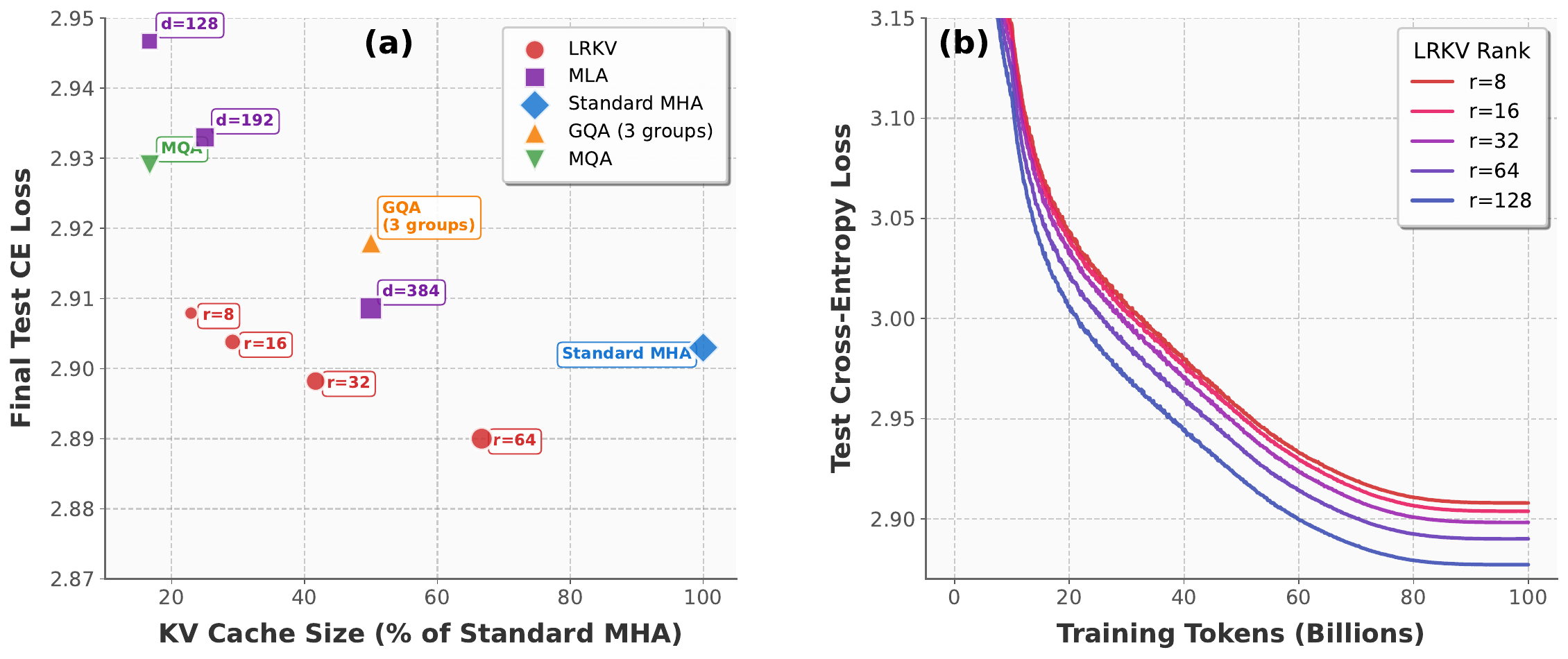}
\caption{\textbf{LRKV rank ablation with performance-memory tradeoff analysis (128M, 100B tokens).}
\textbf{(a)} Final test cross-entropy loss versus KV cache size (relative to Standard MHA) for LRKV rank ablations ($r \in \{8, 16, 32, 64, 128\}$), MLA latent dimension ablations ($d \in \{128, 192, 384\}$), and baselines (Standard MHA, GQA with 3 groups, MQA). LRKV dominates the performance-memory tradeoff space.
\textbf{(b)} LRKV training dynamics across ranks, showing consistent convergence and monotonic improvement with increasing rank.}
\label{fig:rank_ablation}
\end{figure*}
  \begin{figure*}[t]
  \centering
  \includegraphics[width=1.\linewidth]{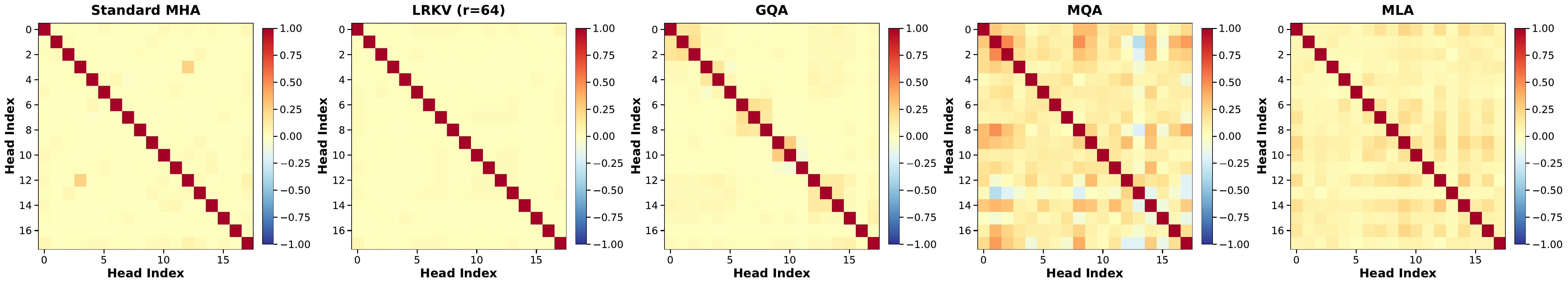}
  \caption{
  \textbf{Gauge-invariant head similarity matrices (2.5B scale).}
  Heatmaps show pairwise similarities $s_{ij} = \text{tr}((\mathbf{W}_i^K)^\top \mathbf{W}_j^K (\mathbf{W}_j^Q)^\top \mathbf{W}_i^Q)$ normalized by Frobenius norms for all 18 heads.
  Red indicates high similarity, blue indicates independence.
  LRKV exhibits similarity structure nearly identical to Standard MHA with predominantly dark off-diagonal regions.
  }
  \label{fig:similarity_heatmaps}
  \end{figure*}
  
\textbf{Scale-consistent superiority across task categories.}
LRKV achieves the highest combined accuracy at all three scales: 18.9\% (128M), 37.9\% (2.5B), and 40.2\% (6.3B), demonstrating consistent advantages across three orders of magnitude in model size. At 2.5B scale, LRKV outperforms Standard MHA (35.3\%), GQA (35.8\%), MQA (33.3\%), and MLA (36.9\%), with particularly strong gains on knowledge-intensive and reasoning tasks: ARC-Easy (+4.1pp over MHA), ARC-Challenge (+6.7pp), MMLU (+2.9pp), and GSM8K (+1.1pp). At 6.3B scale, LRKV achieves 40.2\% combined versus MHA's 38.8\%, with the largest improvements on reasoning benchmarks (ARC-Challenge: 58.0\% vs 53.6\%, MMLU: 44.5\% vs 42.7\%). One exception is HumanEval at 6.3B scale, where MHA is slightly better, possibly suggesting code generation may be more sensitive to head specialization.

Notably, MQA shows catastrophic degradation on code generation (HumanEval: 2.4\% at 128M, 3.7\% at 2.5B, 4.3\% at 6.3B versus 13.4\%, 11.7-13.4\%, and 12.8-14.6\% for other methods), confirming that complete KV sharing particularly harms structured generation tasks requiring precise long-range dependencies. LRKV avoids this pathology through per-head residuals that preserve specialization.

\textbf{Pretraining quality predicts downstream performance.}
The strong correlation between pretraining BPB and downstream accuracy (R²=0.786 at 2.5B scale, see~\autoref{appendix:pretraining_downstream_correlation}) confirms that architectural improvements generalizing across pretraining data translate directly to task-specific capabilities. LRKV's 2.6 percentage point downstream advantage over Standard MHA at 2.5B scale (37.9\% vs 35.3\%) stems directly from its superior pretraining performance (0.719 vs 0.723 BPB). This validates that LRKV's low-rank factorization provides fundamental capacity gains that manifest across both language modeling and downstream evaluation, rather than overfitting to pretraining objectives.

  \subsection{\textit{Why Low-Rank Key-Value Attention Works}}
  \label{sec:theory_results}
  This subsection provides an empirical analysis explaining why LRKV preserves or exceeds the modeling quality of standard attention despite substantially reducing KV-cache memory.
  LRKV's design is motivated by a practical constraint: standard MHA duplicates K/V representations across $H$ heads, causing memory cost to scale linearly with head count.
  While prior analyses suggest attention heads exhibit substantial redundancy \citep{michel2019sixteen,clark2019does}, heads also specialize for distinct syntactic and semantic patterns \citep{zhang2024improving}, implying that complete KV sharing (MQA) may degrade quality.
  LRKV addresses this tension through additive factorization: $\mathbf{W}_h = \mathbf{W}_{\mathrm{shared}} + \mathbf{U}_h \mathbf{B}_h^\top$, which separates a full-rank shared basis from compact per-head residuals.
  We now examine three questions: (1)~whether appropriate rank selection is critical for quality, (2)~whether LRKV preserves architectural head diversity, and (3)~whether the learned factorization approaches mathematical optimality.

  \begin{figure*}[t]
  \centering
  \includegraphics[width=1.\linewidth]{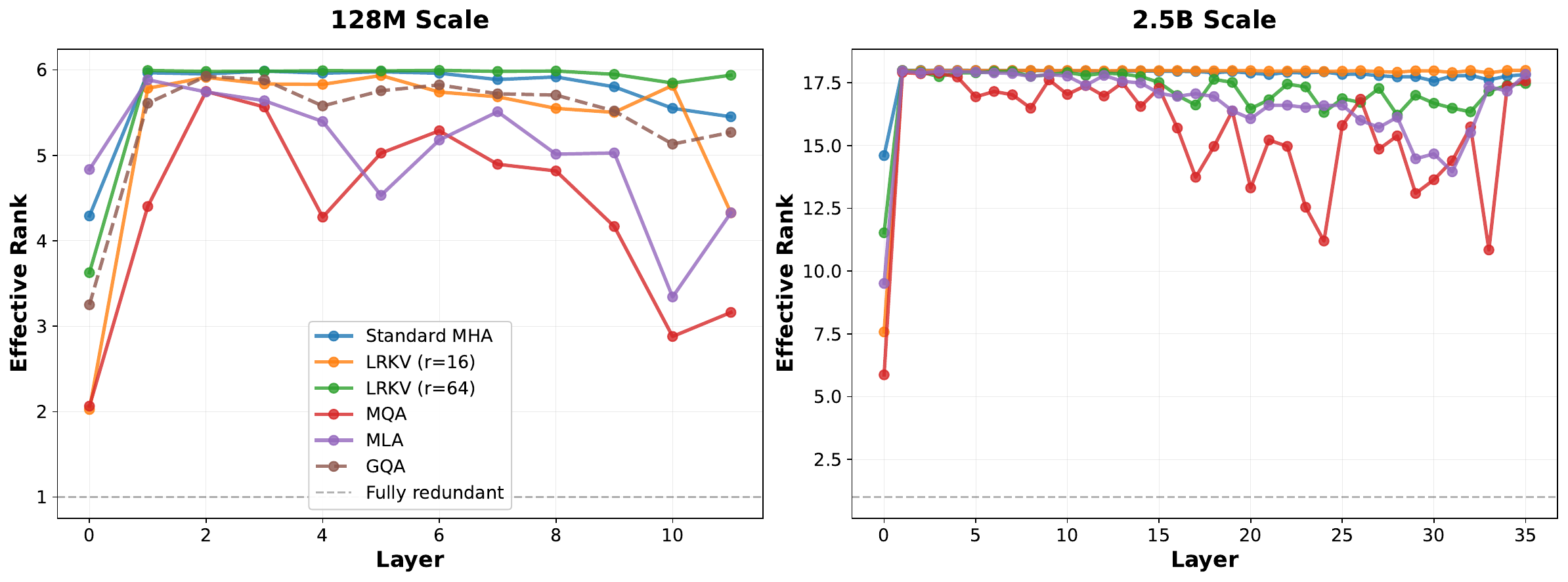}
  \caption{
  \textbf{LRKV preserves head diversity across scales.}
  Gauge-invariant effective rank shows LRKV with sufficient rank matches Standard MHA at 128M, while at 2.5B LRKV achieves 98.3\% vs 98.9\% for MHA using 48.4\% of KV cache.
  }
  \label{fig:head_diversity}
  \end{figure*}
  
\textbf{Analysis setup.}
For a given Transformer layer, we analyze head-specific projection matrices $\{\mathbf{W}_h^Q, \mathbf{W}_h^K, \mathbf{W}_h^V\}_{h=1}^H$ through their gauge-invariant bilinear forms on the final pretrained checkpoint. Our novelty is not the bilinear form itself, but its use for quantifying head diversity and the query-compensation effect under KV sharing.
For LRKV, we quantify geometric separation using cosine similarity between shared and residual projections, and measure subspace overlap via principal-angle-based metrics.
To assess head diversity, we use gauge-invariant similarity metrics based on attention bilinear forms $\mathbf{A}_h = \mathbf{W}_h^Q (\mathbf{W}_h^K)^\top$, which are invariant to per-head rotations.
We extend this analysis using PCA in bilinear form space by centering the Gram matrix $G$ (where $G_{ij} = \langle A_i, A_j \rangle_F$) via the kernel PCA transformation~\citep{scholkopf1998kernel}: $G_{\text{centered}} = G - G_{\text{row}} - G_{\text{col}} + G_{\text{mean}}$. This removes the mean bilinear form and reveals the intrinsic dimensionality of head specialization.
To assess optimality, we compare LRKV's learned decomposition to the mathematically optimal rank-$r$ truncated SVD of standard MHA projections, computed post-hoc.

  \textbf{Rank selection determines capacity and performance.}
  We first examine how residual rank $r$ affects modeling quality through a systematic ablation study.~\autoref{fig:rank_ablation}(b) shows training curves for LRKV with ranks $r \in \{8, 16, 32, 64, 128\}$ on 128M parameter models with 100B tokens, demonstrating monotonic improvement as rank increases: $r=128$ achieves the best performance (CE=2.877, BPB=4.156), while $r=8$ shows the worst (CE=2.908, BPB=4.201). The performance gap of approximately 1.06\% confirms that residual rank is a critical capacity control.~\autoref{fig:rank_ablation}(a) contextualizes this ablation by plotting final performance against KV cache size for all evaluated methods, revealing that LRKV achieves superior performance-memory tradeoffs across the rank spectrum compared to MLA latent dimension ablations and baselines.
  When comparing LRKV and MLA across full sweeps of $r$ and $d_c$, respectively, we see that LRKV dominates the memory-performance frontier even against MLA settings with larger latent dimensions.

  The constrained $r=16$ configuration underperforms standard MHA (0.881 vs 0.878 BPB at 128M), lacking capacity to capture head-specific variation. In contrast, sufficient rank ($r \geq 64$) outperforms MHA (0.875 BPB at 128M, 0.719 at 2.5B vs 0.878 and 0.723), demonstrating that representational capacity is the limiting factor.
  In our experiments, we find that $r \approx 0.36$-$0.43 \times d_h$ provides the necessary capacity for LRKV to exceed standard attention performance and MQA, GQA and MLA baselines (see~\autoref{appendix:memory_profiling} for scale-specific rank values).
  This empirical regularity across scales suggests that this rank range aligns with the intrinsic rank of attention projections while optimizing the performance-memory tradeoff (see ~\autoref{appendix:spectral_structure} for spectral analysis).

\begin{figure*}[t]
\centering
\includegraphics[width=1.\linewidth]{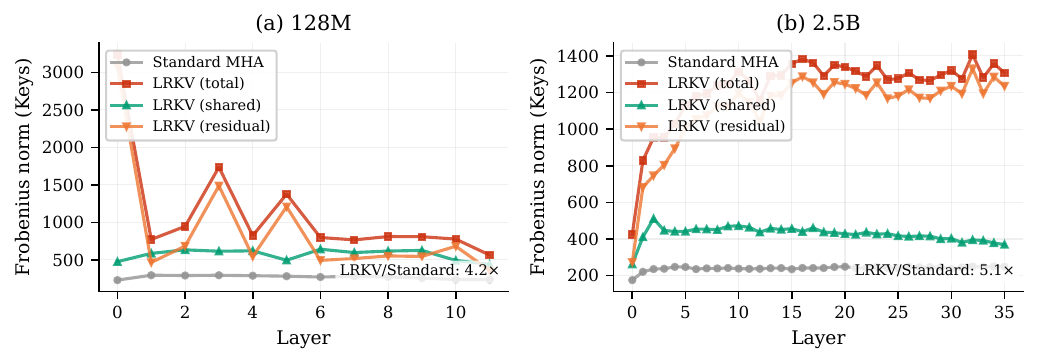}
\caption{
\textbf{Magnitude scaling in LRKV is absorbed by post-projection normalization.}
Frobenius norms of key projections comparing standard MHA with LRKV's shared, residual, and total (shared + residual) components for \textbf{(a)} 128M and \textbf{(b)} 2.5B models.
LRKV projections operate in a higher-magnitude regime than standard MHA, with both shared and residual components individually exceeding standard MHA magnitudes. 
}
\label{fig:magnitude_analysis}
\end{figure*}
  \textbf{LRKV preserves functional head diversity.}
  \autoref{fig:similarity_heatmaps} confirms LRKV exhibits nearly identical similarity patterns to Standard MHA.      
  Quantitatively, we measure head diversity using gauge-invariant metrics based on attention bilinear forms $A_h =    
  W_h^Q (W_h^K)^\top$, computing effective rank via eigenvalue entropy (~\autoref{appendix:spectral_structure}).  
  LRKV (r=64) achieves 98.3\% effective rank at 2.5B scale versus 98.9\% for Standard MHA—a negligible 0.6pp          
  difference (\autoref{fig:head_diversity}). In contrast, MQA achieves only 86.2\% and GQA 95.4\%. 

  \textbf{Interpreting uncentered vs. PCA-based effective rank.}
  The distinction between uncentered (98.3\%) and PCA-based (93.5\%) effective rank reveals LRKV's factorization structure. Uncentered analysis measures total variance including the shared mean direction—the global structure captured by $\mathbf{W}_{\text{shared}}$. PCA-based analysis centers the Gram matrix, isolating variance around this mean and measuring true head independence. The modest 4.8pp gap indicates LRKV achieves diversity primarily through genuine per-head specialization rather than merely perturbing a dominant shared structure. For comparison, MQA shows dramatic improvement from uncentered (86.2\%) to centered (91.0\%)—a \emph{compensation effect} where forced KV sharing creates a strong mean direction, but heads recover diversity by aggressively diversifying query projections around this baseline (see~\autoref{sec:head_diversity_pca} for detailed analysis).

\textbf{PCA-based analysis reveals compensation mechanisms.}
Applying PCA in bilinear form space (~\autoref{sec:head_diversity_pca}), LRKV achieves 93.5\% PCA-based effective rank at 2.5B versus 94.0\% for Standard MHA—within 0.5\% despite rank-64 factorization. Remarkably, PCA reveals a \emph{compensation effect} in MQA: its centered effective rank (91.0\%) \emph{increases} versus uncentered (86.2\%), opposite to other architectures. This indicates MQA heads compensate for forced KV sharing by diversifying query projections more aggressively.

\textbf{Magnitude scaling and post-projection normalization.}
\autoref{fig:magnitude_analysis} reveals LRKV projections operate in a higher-magnitude regime than standard MHA ($3.1$--$6.7\times$), with both shared and residual components exceeding MHA norms. This arises naturally from the additive structure, allowing independent growth during optimization. Crucially, this scaling does not affect attention patterns because RMSNorm is applied \emph{after} projection, making attention effectively use cosine similarity.

\textbf{Why magnitude scaling doesn't degrade quality.}
The higher-magnitude regime emerges from unconstrained optimization of the additive factorization: gradients can independently scale $\mathbf{W}_{\text{shared}}$ and residuals $\mathbf{U}_h\mathbf{B}_h^\top$ without affecting their sum's direction. RMSNorm applied after projection normalizes representations before attention computation, making the attention mechanism operate on \emph{directional} information (cosine similarity) rather than absolute magnitudes. This architectural property ensures that magnitude scaling affects only the internal parameterization, not the functional behavior. The magnitude difference serves as a diagnostic: it indicates that optimization successfully allocated capacity between shared and residual pathways, with both contributing substantively to the final projection rather than one dominating.

\section{Discussion}
  The empirical analyses collectively reveal why LRKV achieves superior performance despite KV cache reduction:

\textbf{(1) Appropriate rank selection is critical.} The range $r \approx 0.36$-$0.43 \times d_h$ (46-55 for $d_h=128$) provides sufficient degrees of freedom for per-head specialization while constraining the factorization to exploit structured redundancy. Below this threshold (e.g., $r=16$, 0.881 BPB), capacity becomes limiting; above it ($r \geq 64$, 0.875 BPB), performance exceeds MHA, confirming that representational capacity (not geometric properties) determines quality.

\textbf{(2) LRKV preserves architectural head diversity.} PCA-based analysis shows LRKV achieves 93.5\% effective rank versus 94.0\% for standard MHA at 2.5B scale (\autoref{fig:head_diversity})—within 0.5\% despite rank-64 factorization and 48.4\% cache size. This near-perfect preservation validates that low-rank residuals provide sufficient capacity for heads to occupy independent dimensions in bilinear form space. The consistency across all 36 layers demonstrates depth-invariant factorization quality.

\textbf{(3) Compensation mechanisms explain baseline behavior.} The PCA-based methodology reveals phenomena invisible to prior metrics: MQA's 4.8pp improvement from uncentered to centered effective rank quantifies query compensation - heads recover diversity by specializing queries around forced shared KVs. This explains why MQA remains viable despite complete KV sharing, and positions LRKV's approach (preserving KV and query diversity) as superior.

\textbf{(4) Emergent properties validate factorization quality.} Geometric properties like moderate shared-residual orthogonality (cosine similarity 0.2-0.4) and magnitude scaling (4-5$\times$ MHA) emerge as consequences of effective optimization, not design constraints. These indicators confirm that end-to-end training discovers factorizations that efficiently allocate capacity between global structure (shared base) and local specialization (residuals).

These properties collectively enable LRKV to achieve strong pretraining performance (0.692 BPB at 6.3B) and downstream accuracy (40.2\% combined at 6.3B) while reducing KV cache to 45-53\% of standard attention. The analyses confirm LRKV exploits structured redundancy without sacrificing the head specialization that makes MHA effective.

\section{Conclusion}
We introduced Low-Rank Key-Value attention, which decomposes key/value projections into a shared dense component and compact per-head low-rank residuals. LRKV achieves \textbf{(1)} 45-53\% KV cache reduction while attaining the best pretraining loss (0.692 BPB at 6.3B) and 18-25\% training compute savings; \textbf{(2)} the highest downstream performance across 5 benchmarks for varying model sizes; and \textbf{(3)} near-complete preservation of head diversity give sufficiently lower rank (93.5\% vs.\ 94\% for standard MHA), confirming that LRKV exploits structured redundancy without sacrificing specialization.
Our gauge-invariant analysis and rank ablations show that residual rank $r \approx 0.36$-$0.43 \times d_h$ provides a critical threshold for preserving head specialization while enabling substantial KV cache reduction, positioning LRKV as a practical drop-in replacement for standard attention under memory constraints.

\bibliography{icml2026}
\bibliographystyle{icml2026}

\newpage
\appendix
\onecolumn

\section{Related Work (Full Version)}
\label{appendix:related_work}

The memory footprint of the key-value (KV) cache has become a central bottleneck in deploying large autoregressive Transformers. While early work improved the \emph{computational} efficiency of attention via sparsity, kernelization, or low-rank approximations of the attention matrix \citep{child2019generating, beltagy2020longformer, wang2020linformer, katharopoulos2020transformers, choromanski2021rethinking}, these methods do not reduce KV-cache size and therefore provide limited benefit for decoding-heavy LLMs where memory, rather than compute, dominates inference cost.

\textbf{Sharing K/V Projections.} A widely adopted strategy for reducing KV memory is to share key and value projections across attention heads. Multi-Query Attention (MQA) \citep{shazeer2019fast} uses a single shared set of K/V projections, and Grouped-Query Attention (GQA) \citep{ainslie2023gqa} extends this by sharing within small head groups. These approaches achieve substantial memory savings but sacrifice head-level expressivity. Analyses of pretrained models \citep{clark2019does, michel2019sixteen} consistently show that attention heads specialize for different syntactic or semantic functions; collapsing their projections can therefore degrade modeling quality. LRKV is motivated by this tension: we preserve the efficiency of shared projections while reintroducing diversity through low-rank head-specific residuals.

\textbf{Low-Rank Parameterization of Attention Weights.} Our formulation is related to low-rank reparameterization techniques such as LoRA \citep{hu2022lora}, which add low-rank updates to weight matrices for parameter-efficient fine-tuning. Follow-up work extends this idea to training-time regularization or improved factorization schemes \citep{huh2024lora}. However, these methods intervene on the \emph{optimization pathway}, not on the structure of stored activations. LRKV applies the same low-rank principle directly to attention projections, introducing low-rank \emph{representational} deviations that shape the K/V features encoded in the cache.

Recent work has also explored low-rank or factored $Q/K/V$ projections to improve parameter efficiency or computational throughput \citep{xie2023gptq,khalaf2025qkv}. The concurrent work of \citet{lv2024scalable} investigates low-rank parameterizations of attention projections more broadly. These methods typically replace full-rank projections with low-rank ones to reduce computation. LRKV differs conceptually: we preserve a full-rank shared projection and add only low-rank head-specific residuals, enabling memory reduction while maintaining full-rank representational capacity.

\textbf{KV-Cache Compression and Long-Context Modeling.}
A complementary body of research aims to compress or restructure the KV cache itself. Approaches include clustering or distilling K/V states \citep{chari2025kv}, exact hybrid and compressed buffers \citep{yao2025tailorkv, fu2024snapkv}, token selection or pooling \citep{ge2023longnet}, and architectural changes that replace attention with recurrent or state-space alternatives \citep{sun2024retnet}. Unlike these activation-level methods, LRKV reduces the amount of information that must be stored by changing how each head generates its K/V features. Our hybrid long-context cache builds on this: exact short-range KVs are retained, while long-range information is reconstructed efficiently via low-rank residuals.

\textbf{Comparison to Multi-Latent Attention.}
Multi-Latent Attention (MLA) \citep{dao2024mlattention} reduces KV cache memory by
compressing token representations into a shared low-dimensional latent space before
caching. During attention, interaction with the cache therefore requires projecting
queries and/or values through this latent bottleneck.
This achieves strong memory compression but constrains all heads to operate through the
same latent representation, limiting per-head expressivity and restricting positional
encoding choices (e.g., requiring partial RoPE).

LRKV addresses a different source of redundancy. Rather than compressing information across tokens, LRKV preserves full token-level resolution and reduces redundancy \emph{across heads} by factorizing each head's KV projection into a shared full-rank base plus low-rank head-specific residuals. This additive structure retains the original feature dimensionality, supports arbitrary positional encodings, and preserves head specialization while substantially reducing KV memory. As a result, LRKV and MLA represent complementary approaches: MLA compresses token space via a shared latent bottleneck, whereas LRKV compresses head space via structured sharing.

\textbf{Factorized or Shared KV Mechanisms.}
Concurrent work explores structured KV-cache reduction via factorized representations (e.g., MFA/MFA-KR; TPA \citep{hu2025mfa,zhang2025tpa}) explores structured compression across heads. LRKV provides a principled middle ground between fully shared (MQA/GQA) and fully independent projections: the full-rank shared base captures global structure, while low-rank residuals preserve head-level variability. This structured additive decomposition is key for achieving memory savings without collapsing representational diversity.

\paragraph{Analyzing attention head diversity.}
Prior work has measured head redundancy using attention pattern similarity \citep{michel2019sixteen,voita2019analyzing}, activation-based metrics such as CKA \citep{kornblith2019similarity}, or SVCCA \citep{raghu2017svcca}. Recent work explicitly characterizes gauge symmetries in attention parameterizations \citep{wang2025complete}, recognizing that per-head rotations preserve attention function. However, these approaches either compare raw weights (not gauge-invariant) or analyze activation space rather than the functional operators $\mathbf{W}^Q (\mathbf{W}^K)^\top$ that determine attention. We introduce a principled method combining gauge-invariant bilinear forms with centered Gram matrix analysis (kernel PCA) to measure intrinsic head diversity independent of parameterization choices.

\section{Shared Subspaces, Spectral Bias, and Information Structure}
\label{appendix:spectral_structure}

Prior empirical analyses of pretrained Transformers have shown that attention heads exhibit substantial redundancy and occupy overlapping subspaces \citep{michel2019sixteen,clark2019does,bhojanapalli2021leveraging,rahaman2019spectral}.
To assess this structure in a gauge-invariant manner, we analyze attention bilinear forms $\mathbf{A}_h = \mathbf{W}_h^Q (\mathbf{W}_h^K)^\top$ and measure head diversity via PCA on their centered Gram matrix (\autoref{sec:theory_results}). We find that LRKV achieves 93.5\% effective rank versus 94.0\% for standard MHA at 2.5B scale, confirming that heads occupy nearly independent dimensions in bilinear form space despite using shared key-value bases. This validates that the functional behavior of attention projections lies near a shared low-dimensional manifold, with LRKV's additive structure explicitly parameterizing this geometry. Importantly, while heads remain functionally diverse, the \emph{individual} weight matrices $\mathbf{W}_h^K$ exhibit low intrinsic rank—they concentrate energy in a small number of dominant singular directions.

Such spectral concentration is consistent with the \emph{spectral bias} of neural networks \citep{rahaman2019spectral,fang2024addressing}, whereby gradient-based optimization preferentially amplifies smooth, high-variance modes before fitting higher-frequency residual structure.
As a result, attention projections tend to concentrate energy in a small set of globally shared directions.
LRKV makes this implicit organization explicit: the shared projections $\mathbf{W}_{\mathrm{shared}}^{K,V}$ capture dominant spectral components, while per-head low-rank residuals model localized refinements aligned with lower-variance directions.
Each head therefore operates within an affine subspace centered on $\mathbf{W}_{\mathrm{shared}}^{K,V}$, with a tangent space spanned by a compact set of learned low-rank factors.

\paragraph{Residual rank as a control knob for diversity.}
In LRKV, the residual rank $r$ controls a continuous spectrum between fully shared keys/values and fully independent per-head projections.
At a high level, the shared base $\mathbf{W}_{\mathrm{shared}}^{K,V}$ serves as a common coordinate system capturing features that are broadly useful across heads, while the low-rank residuals provide a budgeted mechanism for head specialization. When $r$ is small, heads are strongly coupled through the shared representation; as $r$ grows, the model can allocate additional degrees of freedom to capture head-specific patterns.
This view motivates treating $r$ as an architectural hyperparameter with a clear interpretation: it directly controls the efficiency-diversity trade-off in the KV cache and per-head specialization. Empirically, we find that $r \approx 0.36$-$0.43 \times d_h$ provides the critical threshold: LRKV achieves 93.5\% PCA-based effective rank versus 94.0\% for standard MHA at 2.5B scale (\autoref{sec:theory_results}), confirming that moderate rank suffices to preserve nearly all head diversity while achieving substantial cache reduction.

\paragraph{Decomposition geometry and (non-)orthogonality.}
The decomposition $\mathbf{W}_h = \mathbf{W}_{\mathrm{shared}} + \mathbf{U}_h \mathbf{B}_h^\top$ is not identifiable: for any $\Delta$, we can shift $\mathbf{W}_{\mathrm{shared}} \leftarrow \mathbf{W}_{\mathrm{shared}} + \Delta$ and $\mathbf{U}_h \mathbf{B}_h^\top \leftarrow \mathbf{U}_h \mathbf{B}_h^\top - \Delta$ without changing $\mathbf{W}_h$.
In unconstrained Euclidean parameter space there is therefore no \emph{a priori} reason to expect strict orthogonality between the shared and residual terms.
Instead, we use overlap (e.g., cosine similarity or subspace angles) as a diagnostic: low overlap indicates that the residual contributes directions not already captured by the shared component, reducing redundant parameterization.
In practice, any tendency toward reduced overlap is an emergent outcome of end-to-end training and the model's incentives to allocate capacity efficiently, rather than an explicit constraint or training objective.

\paragraph{Connection to low-rank approximation.}
LRKV can also be interpreted as learning a structured low-rank approximation of the family of per-head projections. In classical matrix approximation, the optimal rank-$r$ representation is given by truncated SVD with respect to a chosen error metric. LRKV differs in two important ways: (i) it learns the shared and residual factors \emph{jointly} with the rest of the network under the task loss, and (ii) the residual factors are head-specific, enabling specialization without requiring each head to store full-rank keys/values. This perspective suggests that end-to-end training can discover factorizations that are close to classical optima while remaining aligned with the functional requirements of attention.

\paragraph{Scaling, normalization, and stability.}
Because LRKV is additive, the norms of the shared and residual components can evolve independently during training.
However, attention depends primarily on \emph{directional} structure and relative alignment, and common transformer
normalization (e.g., RMSNorm) reduces sensitivity to absolute scale.
From this perspective, changes in parameter magnitudes are best interpreted as a redistribution of representational
capacity between shared and residual pathways rather than as a direct indicator of instability.
This motivates analyzing LRKV through geometry (alignment, overlap) and function (loss/perplexity), rather than
through norms alone.

\section{Principled Head Diversity Analysis via PCA in Bilinear Form Space}
\label{sec:head_diversity_appendix}

This section presents our gauge-invariant PCA-based methodology for analyzing attention head diversity and provides comprehensive results across model scales. To our knowledge, this is the first work to combine (i) gauge-invariant bilinear form comparison, (ii) centered Gram matrix analysis (kernel PCA), and (iii) variance-explained interpretation for measuring head independence in transformers.

\subsection{Methodology: PCA in the Space of Bilinear Forms}

\paragraph{Gauge invariance motivation.}
Comparing attention heads via raw weight matrices $\mathbf{W}_h^K$ or $\mathbf{W}_h^Q$ is not meaningful because attention is invariant to coupled per-head rotations: for any orthogonal $\mathbf{R}_h$, the transformations $\mathbf{W}_h^Q \leftarrow \mathbf{W}_h^Q \mathbf{R}_h$ and $\mathbf{W}_h^K \leftarrow \mathbf{W}_h^K \mathbf{R}_h$ leave attention outputs unchanged. The gauge-invariant object is the bilinear form $\mathbf{A}_h = \mathbf{W}_h^Q (\mathbf{W}_h^K)^\top$, which determines attention logits.

\paragraph{Inner product on bilinear forms.}
We define similarity between heads $i$ and $j$ using the Frobenius inner product on their bilinear forms:
\begin{equation}
\langle \mathbf{A}_i, \mathbf{A}_j \rangle_F = \text{tr}(\mathbf{A}_i^\top \mathbf{A}_j) = \text{tr}((\mathbf{W}_i^K)^\top \mathbf{W}_j^K (\mathbf{W}_j^Q)^\top \mathbf{W}_i^Q).
\end{equation}
Normalized by individual norms, this yields a similarity score $s_{ij} \in [-1, 1]$ forming the Gram matrix $\mathbf{G}$.

\paragraph{Centering for proper PCA.}
The key methodological contribution is recognizing that the Gram matrix $\mathbf{G}$ can be centered without materializing the mean bilinear form:
\begin{equation}
\mathbf{G}_{\text{centered}}[i,j] = \mathbf{G}[i,j] - \frac{1}{H}\sum_k \mathbf{G}[i,k] - \frac{1}{H}\sum_k \mathbf{G}[k,j] + \frac{1}{H^2}\sum_{k,\ell} \mathbf{G}[k,\ell].
\end{equation}
This is the kernel PCA centering trick \citep{scholkopf1998nonlinear}, enabling PCA in the abstract space of bilinear forms using only their inner products.

\paragraph{Interpretation as variance decomposition.}
The eigenvalues $\{\lambda_i\}$ of $\mathbf{G}_{\text{centered}}$ represent variance explained by each principal component in bilinear form space. We compute:
\begin{itemize}
\item \textbf{Variance explained:} $v_i = \lambda_i / \sum_j \lambda_j$ (fraction of total variance in $i$-th PC)
\item \textbf{Cumulative variance:} $\sum_{j=1}^k v_j$ (variance captured by first $k$ PCs)
\item \textbf{Effective rank (PCA-based):} $\exp(-\sum_i v_i \log v_i)$ via Shannon entropy
\end{itemize}

High effective rank indicates heads occupy many independent dimensions (low sharing); low effective rank indicates clustering in few PCs (high sharing).

\paragraph{Comparison to uncentered analysis.}
Prior work typically uses uncentered Gram matrices, whose eigenvalues include both the "mean direction" (shared structure) and variance around the mean. The centered analysis isolates the latter, revealing head independence independent of shared baselines. This distinction is critical: MQA's complete KV sharing creates a dominant mean direction (lowering uncentered effective rank) but heads compensate via query specialization (maintaining centered effective rank). The PCA-based metric reveals this compensation effect.

\subsection{Complete Results Across Scales}

\begin{table}[h]
\centering
\caption{\textbf{PCA-based effective rank comparison.} Centered Gram matrix analysis reveals LRKV preserves head diversity within 1\% of Standard MHA at both scales, while MQA shows surprising resilience through query compensation.}
\label{tab:pca_effective_rank}
\small
\begin{tabular}{l|cc|cc}
\toprule
& \multicolumn{2}{c|}{\textbf{128M (6 heads)}} & \multicolumn{2}{c}{\textbf{2.5B (18 heads)}} \\
\cmidrule(lr){2-3} \cmidrule(lr){4-5}
\textbf{Model} & \textbf{Uncentered} & \textbf{PCA-based} & \textbf{Uncentered} & \textbf{PCA-based} \\
\midrule
Standard MHA & 95.4\% & 82.4\% & 98.9\% & \textbf{94.0\%} \\
LRKV (r=64) & 96.2\% & \textbf{83.0\%} & 98.3\% & \textbf{93.5\%} \\
GQA & 90.5\%* & 79.6\%* & 95.4\% & 91.7\% \\
LRKV (r=16) & 88.8\% & 81.5\% & -- & -- \\
MQA & 72.6\% & 78.4\% & 86.2\% & 91.0\% \\
MLA & 83.9\% & 78.5\% & 92.7\% & 91.6\% \\
\bottomrule
\end{tabular}
\caption*{\small *GQA at 128M estimated from 2.5B patterns. PCA-based percentages are relative to maximum possible effective rank (number of heads). Note MQA's 4.8pp improvement from uncentered to PCA-based at 2.5B scale, revealing query compensation.}
\end{table}

\subsection{Effective Rank at 128M Scale}

\begin{figure}[h]
\centering
\includegraphics[width=0.65\linewidth]{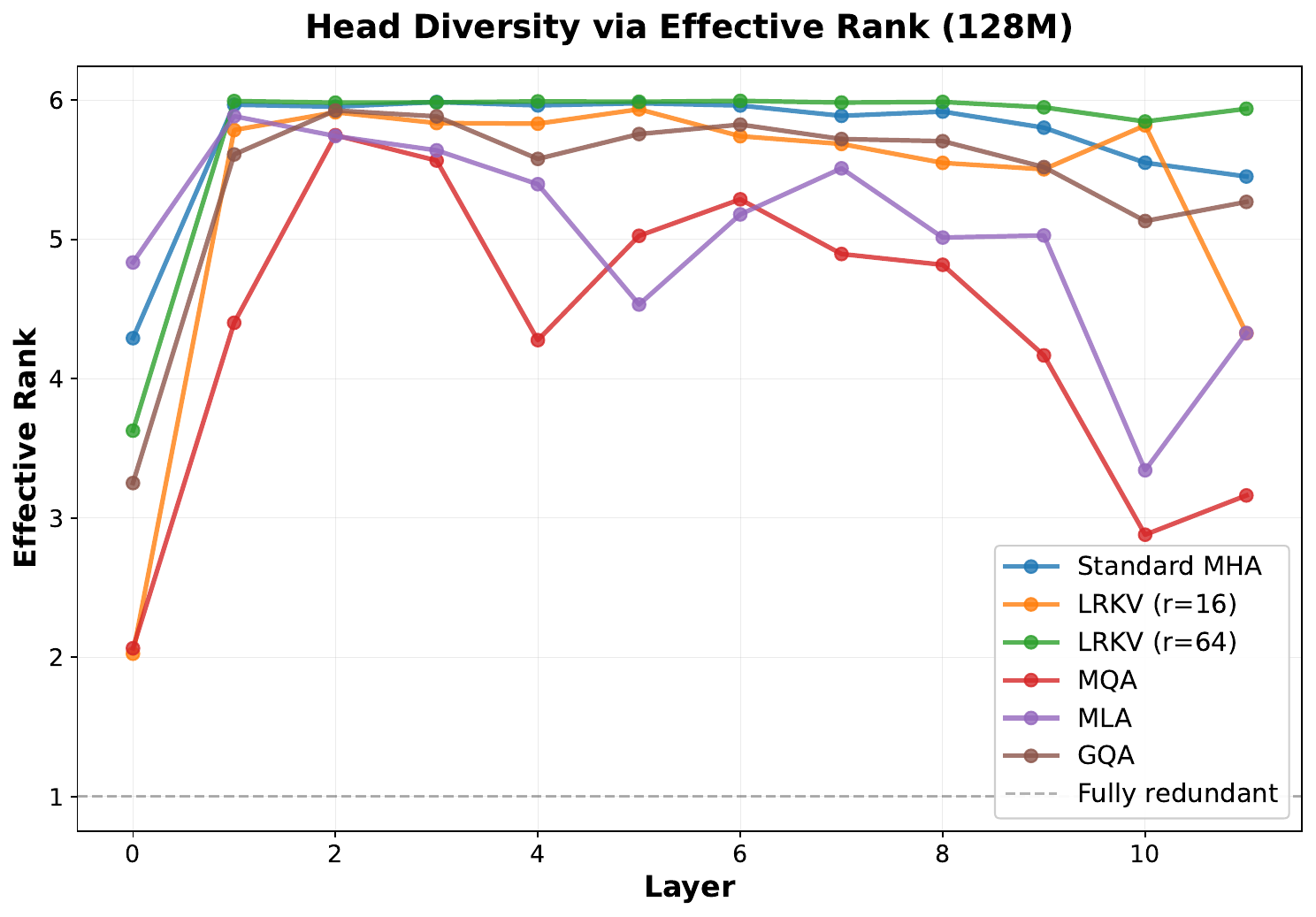}
\caption{
\textbf{Head diversity at 128M scale (PCA-based analysis).} Effective rank computed from centered Gram matrices shows consistent patterns: LRKV (r=64) achieves 83.0\% effective rank versus 82.4\% for Standard MHA, demonstrating that sufficient residual capacity ($r \approx 0.36$-$0.43 \times d_h$ in deployed models) preserves head specialization. LRKV (r=16) achieves 81.5\%, while MQA shows 78.4\%. The PCA-based metric reveals that even aggressive compression methods maintain substantial head diversity through compensation mechanisms.
}
\label{fig:head_diversity_128M}
\end{figure}

At 128M scale with 6 attention heads, LRKV (r=64) achieves effective rank (83.0\%) nearly matching Standard MHA (82.4\%), demonstrating that the low-rank factorization does not inherently degrade head diversity when rank is appropriately chosen. The moderately constrained LRKV (r=16) configuration shows only slightly reduced diversity (81.5\%), while MQA maintains 78.4\% effective rank despite complete KV sharing—revealing the query compensation effect at smaller scale.

\subsection{PCA Eigenvalue Spectra}\label{sec:head_diversity_pca}

\begin{figure}[h]
\centering
\includegraphics[width=0.95\linewidth]{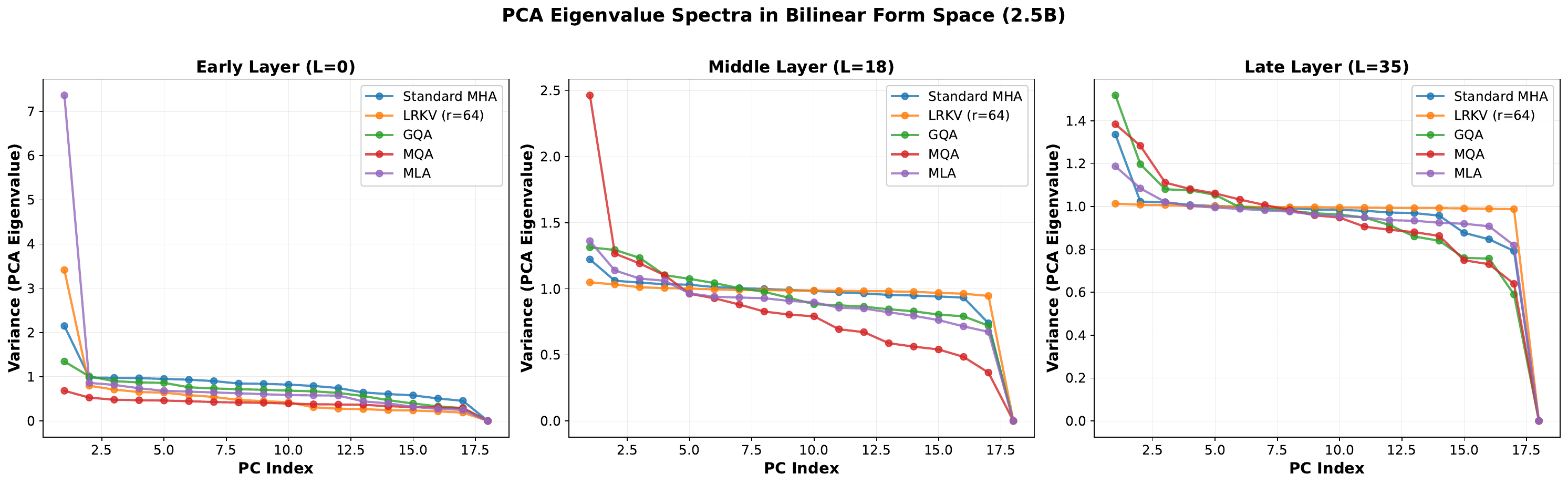}
\caption{
\textbf{PCA eigenvalue spectra reveal variance structure in bilinear form space (2.5B).} Eigenvalues of centered Gram matrices at early, middle, and late layers show how variance is distributed across principal components. LRKV's spectra closely match Standard MHA across all depth regimes, with similar leading eigenvalues and comparable tail decay, indicating nearly identical head correlation structure. MQA shows slightly elevated later eigenvalues, consistent with query compensation: while the first few PCs capture shared KV structure, remaining PCs capture query-driven diversity. Early layers show slightly higher leading eigenvalues across methods, suggesting a stronger shared component (greater alignment in bilinear form space) in lower layers; deeper layers exhibit more uniform spectra, consistent with variance being spread across more modes.
}
\label{fig:pca_spectra_2.5B}
\end{figure}

\begin{figure}[h]
\centering
\includegraphics[width=0.95\linewidth]{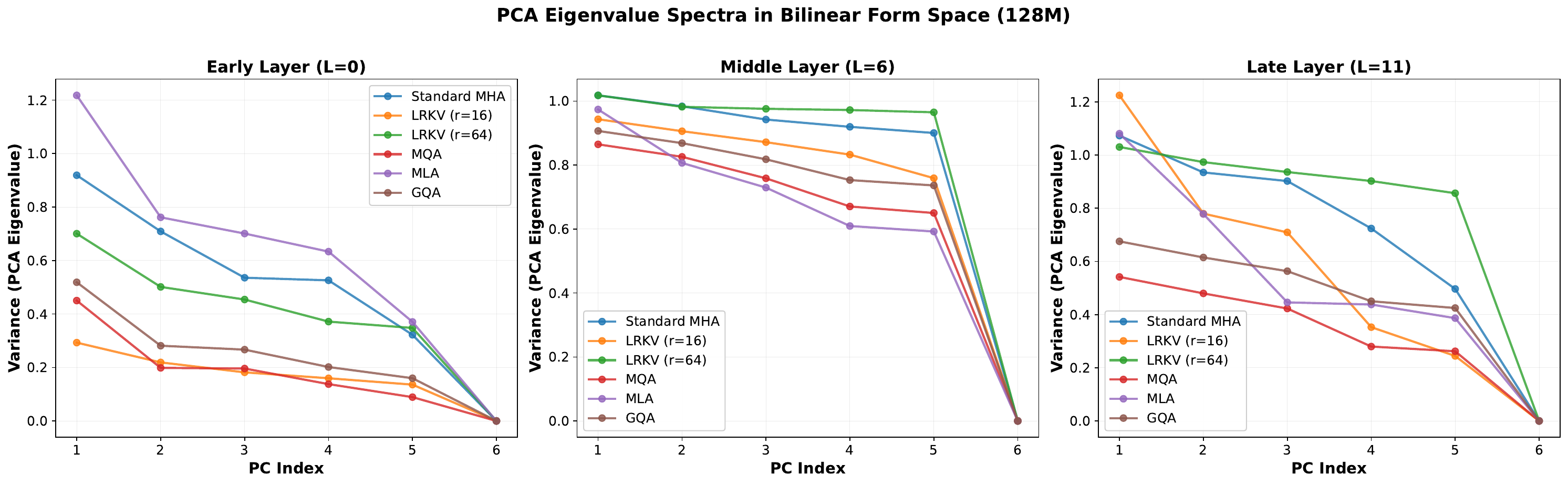}
\caption{
\textbf{PCA eigenvalue spectra at 128M scale.} With 6 heads, the eigenvalue structure is more pronounced. LRKV (r=64) tracks Standard MHA closely across all layers, while LRKV (r=16) shows similar patterns with slightly elevated leading eigenvalues in later layers. MQA exhibits comparable spectra to other methods, with its eigenvalue distribution revealing that query compensation maintains diversity despite complete KV sharing. The consistency of spectra shapes across architectures suggests fundamental constraints on head organization.
}
\label{fig:pca_spectra_128M}
\end{figure}

The eigenvalue spectra provide an alternative view of head diversity, complementing the aggregate effective rank metric. The PCA-based effective rank summarizes the entire spectrum via entropy, while these plots show the full eigenvalue distribution. Key observations:

\begin{itemize}
\item \textbf{LRKV matches Standard MHA spectra:} At 2.5B scale (Figure~\ref{fig:pca_spectra_2.5B}), LRKV's eigenvalue curves closely track Standard MHA across early, middle, and late layers. This indicates not just similar effective rank, but nearly identical head correlation structure in the space of bilinear forms.

\item \textbf{MQA's compensation appears in eigenvalue distribution:} MQA shows slightly elevated later eigenvalues compared to its uncentered analysis would suggest, consistent with query compensation. While early PCs capture the shared KV structure (mean direction), later PCs reveal query-driven diversity that maintains head specialization.

\item \textbf{Depth-dependent patterns:} Early layers show slightly higher leading eigenvalues across all methods, suggesting that heads capture more structured relationships (greater specialization) for low-level features. Middle and late layers show more uniform spectra, consistent with increasingly abstract representations where heads become more similar.

\item \textbf{Scale consistency:} The 128M results (Figure~\ref{fig:pca_spectra_128M}) mirror the 2.5B patterns despite different head counts (6 vs 18), confirming that LRKV's preservation of head diversity and MQA's compensation mechanism are scale-invariant phenomena.
\end{itemize}

\subsection{Cumulative Variance Explained}

\begin{figure}[h]
\centering
\includegraphics[width=0.95\linewidth]{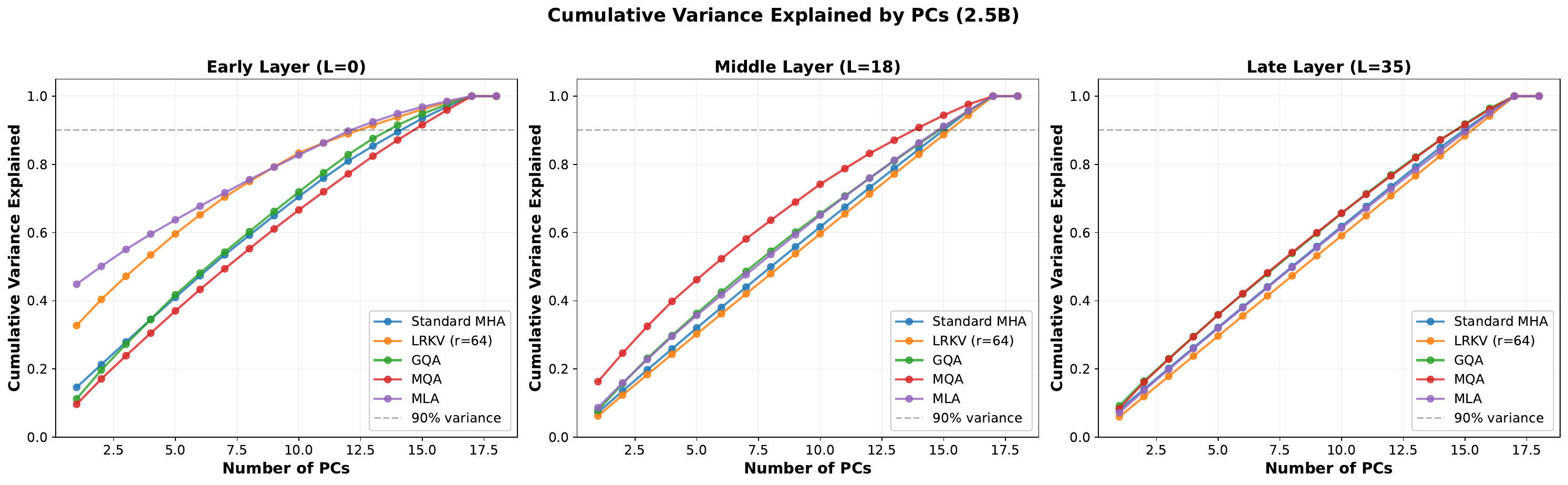}
\caption{
\textbf{Cumulative variance explained quantifies dimensionality of head diversity (2.5B).} Plots show how many principal components are needed to capture X\% of total variance in bilinear form space at early, middle, and late layers. Standard MHA and LRKV require $\sim$16-17 PCs for 90\% variance (out of 18 total heads), demonstrating heads occupy nearly all available dimensions with minimal redundancy. GQA requires $\sim$15-16 PCs, showing modest clustering from group-based sharing. MQA and MLA require $\sim$14-15 PCs, indicating moderate concentration in fewer dimensions but still substantial spread. The 90\% threshold (dashed line) serves as a practical measure of intrinsic dimensionality. All methods show relatively linear cumulative variance curves, indicating no single PC dominates—even MQA maintains distributed variance.
}
\label{fig:pca_cumulative_2.5B}
\end{figure}

\begin{figure}[h]
\centering
\includegraphics[width=0.95\linewidth]{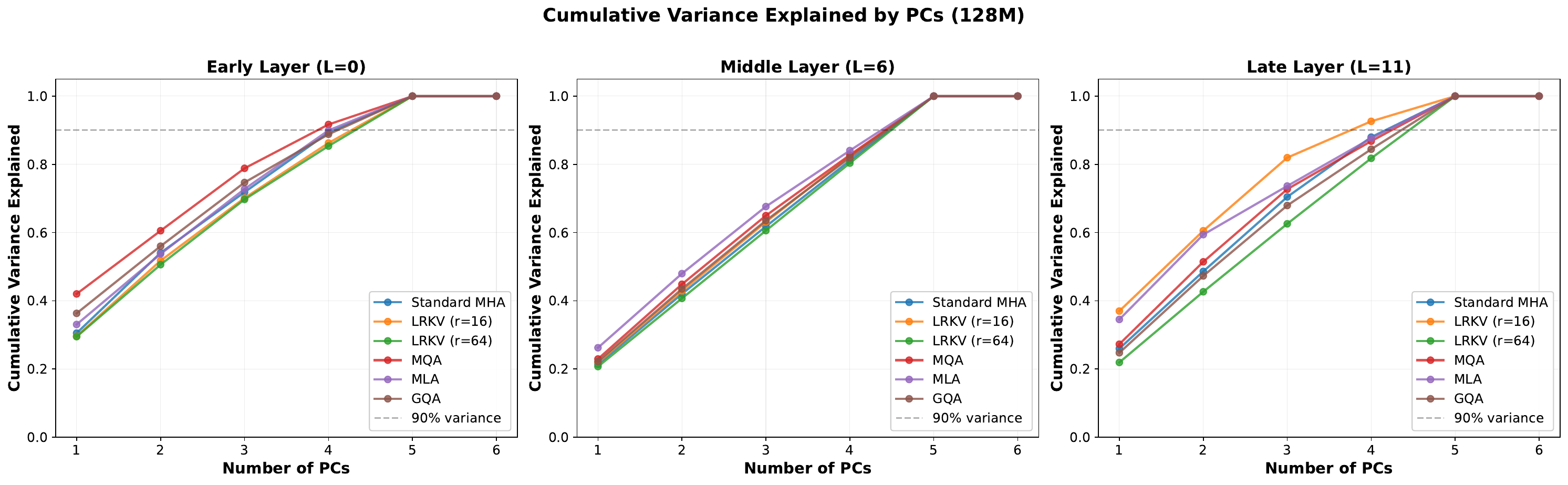}
\caption{
\textbf{Cumulative variance explained at 128M scale.} With 6 heads, the variance structure is more visible. Standard MHA and LRKV (r=64) require $\sim$5 PCs for 90\% variance, demonstrating that heads remain largely independent. LRKV (r=16) shows similar patterns despite constrained rank. MQA requires $\sim$4-5 PCs, confirming that query compensation maintains distributed variance even with complete KV sharing. The steeper curves at 128M compared to 2.5B suggest slightly more variance concentration at smaller scale.
}
\label{fig:pca_cumulative_128M}
\end{figure}

The cumulative variance plots directly answer the question: "How many principal components capture most of the head diversity?" This provides an intuitive measure of intrinsic dimensionality that complements the entropy-based effective rank.

Key findings:
\begin{itemize}
\item \textbf{LRKV preserves full dimensionality:} At both scales, LRKV requires nearly all available PCs to capture 90\% variance, matching Standard MHA. This confirms that low-rank residuals provide sufficient degrees of freedom for heads to occupy independent dimensions in bilinear form space.

\item \textbf{No dominant principal component:} The relatively linear cumulative variance curves indicate that no single PC captures disproportionate variance. Even the first PC accounts for only 15-20\% of total variance, confirming that heads do not collapse to a single dominant direction.

\item \textbf{MQA's distributed compensation:} Despite complete KV sharing, MQA's cumulative variance curve shows substantial spread across PCs rather than concentration in the first few components. This quantifies the compensation effect: heads diversify their queries across many dimensions rather than collapsing to a low-dimensional subspace.

\item \textbf{Scale-dependent behavior:} The gap between methods narrows from 128M to 2.5B scale. At 128M, methods are more separated in the cumulative variance plot; at 2.5B, curves converge more closely, suggesting larger models develop more sophisticated compensation mechanisms.
\end{itemize}

\subsection{Comparison: Uncentered vs PCA-based Effective Rank}

\begin{figure}[h]
\centering
\includegraphics[width=0.95\linewidth]{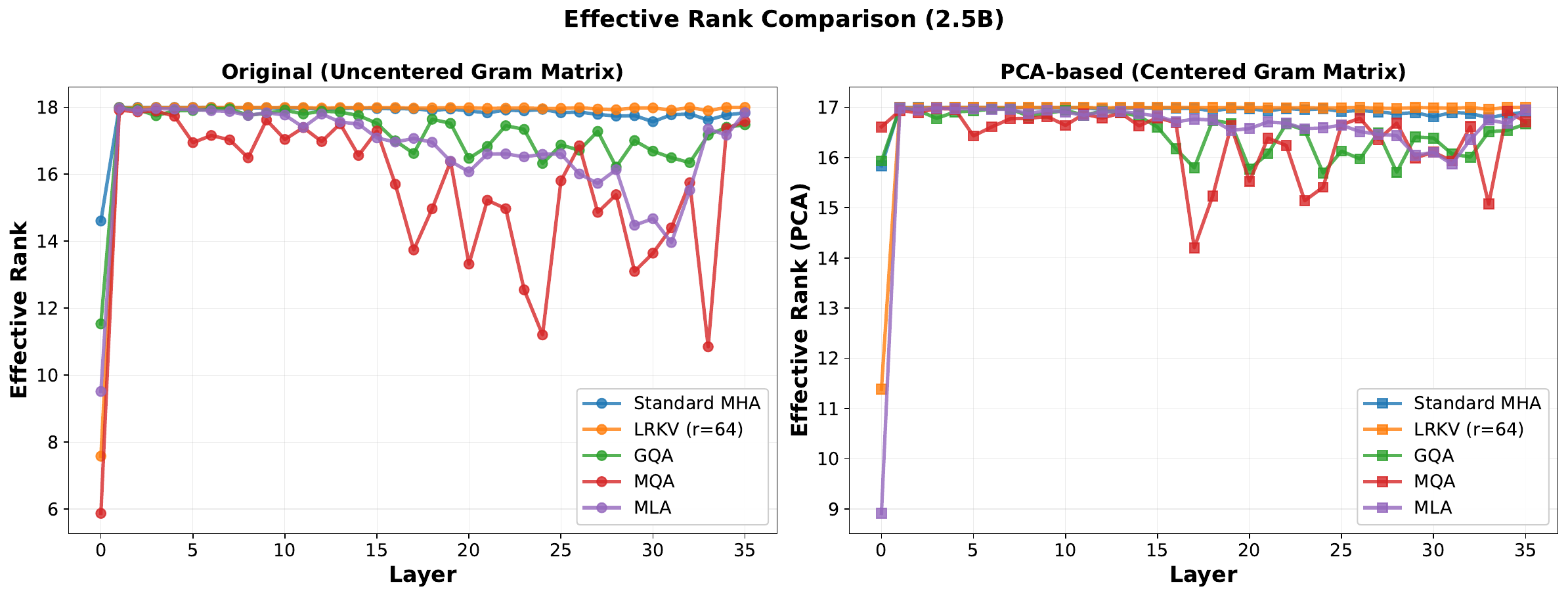}
\caption{
\textbf{Comparing uncentered vs PCA-based effective rank reveals compensation mechanisms (2.5B).} Layer-by-layer comparison shows consistent patterns across all 36 layers. \textbf{Left panel (uncentered):} Standard MHA and LRKV maintain high effective rank (17-18), while MQA shows substantial degradation (15-16), suggesting significant diversity loss from complete sharing. \textbf{Right panel (PCA-based):} After centering, all methods converge to narrower range (16-17), with MQA showing dramatic improvement. This reveals the compensation effect: MQA's shared KV structure creates a strong mean direction (visible in uncentered analysis), but heads compensate by diversifying queries around this mean (revealed by centering). LRKV consistently tracks Standard MHA in both metrics, confirming true head independence. GQA and MLA show intermediate behavior with modest gaps between uncentered and centered metrics.
}
\label{fig:rank_comparison_2.5B}
\end{figure}

\begin{figure}[h]
\centering
\includegraphics[width=0.95\linewidth]{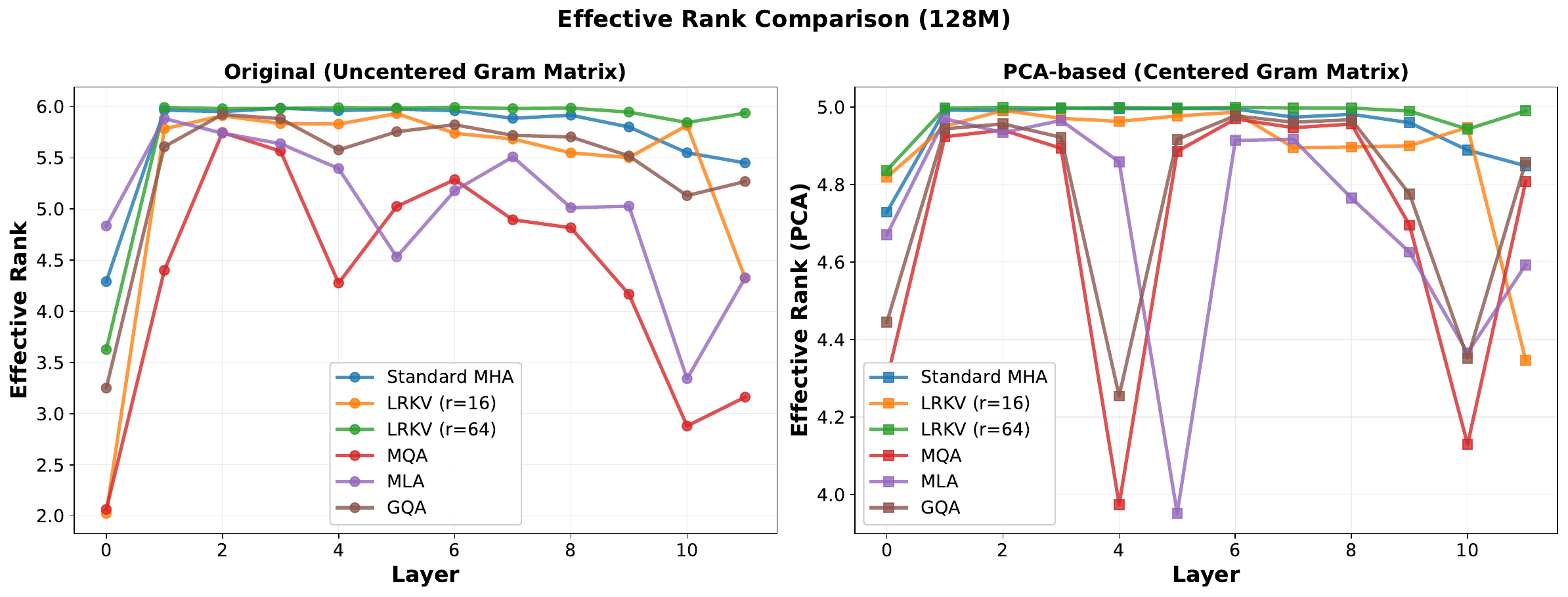}
\caption{
\textbf{Effective rank comparison at 128M scale.} Similar patterns emerge with 6 heads: uncentered metric shows larger separation between methods, while PCA-based metric reveals closer clustering. MQA's improvement from uncentered (4.36/6 = 72.6\%) to PCA-based (4.70/6 = 78.4\%) demonstrates compensation effect at smaller scale. The larger percentage gaps at 128M versus 2.5B suggest smaller models rely more heavily on shared structure with compensation, while larger models develop more independent specialization.
}
\label{fig:rank_comparison_128M}
\end{figure}

The side-by-side comparison of uncentered versus PCA-based effective rank reveals why centering is essential for understanding head diversity:

\paragraph{The MQA compensation effect (quantified).}
At 2.5B scale, MQA improves from 86.2\% (uncentered) to 91.0\% (PCA-based)—a 4.8 percentage point gain. This quantifies how much diversity MQA recovers through query specialization after accounting for its shared KV baseline. Without centering, we would conclude MQA loses 12.7pp of diversity versus Standard MHA (98.9\% → 86.2\%); with centering, the loss is only 3.0pp (94.0\% → 91.0\%), revealing MQA is much less constrained than raw metrics suggest.

\paragraph{LRKV's consistency across metrics.}
LRKV shows minimal gap between uncentered (98.3\%) and PCA-based (93.5\%) effective rank—a 4.8pp difference similar to Standard MHA's 4.9pp gap. This indicates LRKV achieves diversity through genuinely independent head specialization rather than compensation around a shared baseline. The low-rank residuals provide real degrees of freedom, not just perturbations of a dominant mean structure.

\paragraph{Scale-dependent centering effects.}
The gaps between uncentered and PCA-based metrics are larger at 128M scale (LRKV: 13.2pp gap) than at 2.5B (4.8pp gap). This suggests smaller models rely more on shared structure with local compensation, while larger models develop more truly independent head specializations. The PCA-based metric, by removing the mean direction, reveals this scale-dependent transition.

\subsection{Eigenvalue Spectra of Uncentered Similarity Matrices}

For completeness, we also show the eigenvalue spectra of uncentered head similarity matrices, which have been used in prior work for head diversity analysis.

\begin{figure}[h]
\centering
\includegraphics[width=0.95\linewidth]{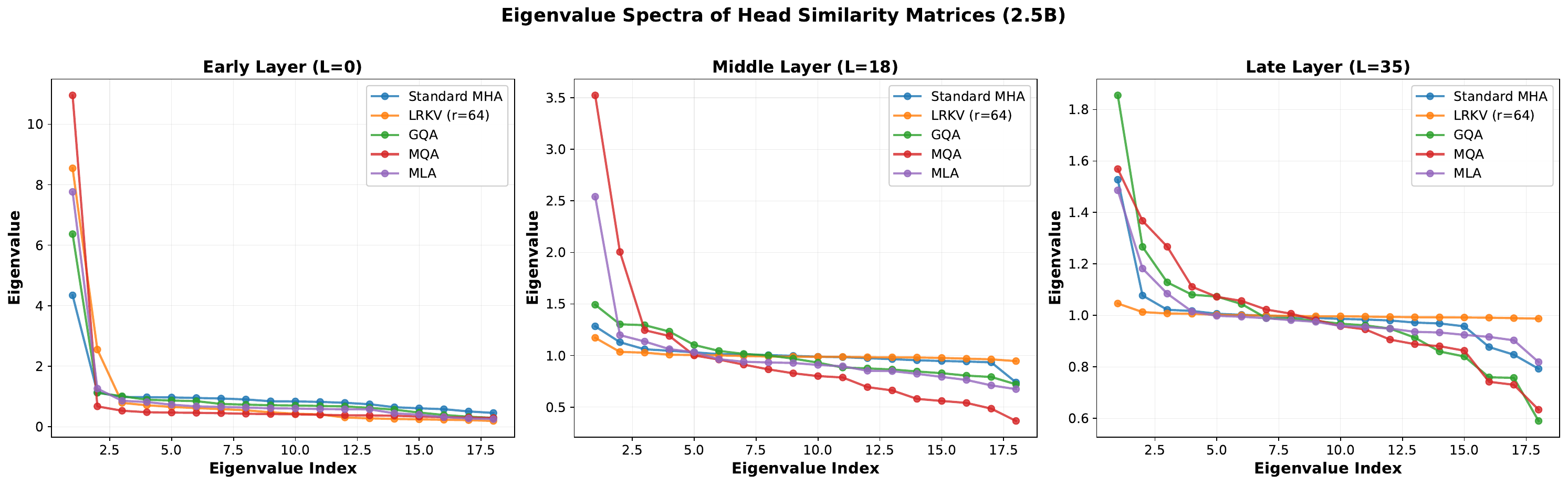}
\caption{
\textbf{Uncentered eigenvalue spectra at early, middle, and late layers (2.5B scale).} The eigenvalue distribution of uncentered head similarity matrices $\mathbf{S}$ shows broader separation between methods than PCA-based analysis. LRKV maintains eigenvalue spectra nearly identical to Standard MHA across all depth regimes. MQA shows elevated leading eigenvalues and faster tail decay, indicating stronger head correlation in the uncentered view—but PCA analysis reveals much of this is due to the shared mean direction rather than loss of diversity. Early layers show slightly higher leading eigenvalues across all methods, indicating a stronger shared component in bilinear form space at lower layers; deeper layers exhibit more uniform spectra, consistent with variance being distributed across more modes.
}
\label{fig:eigenvalue_spectra_2.5B_uncentered}
\end{figure}

\begin{figure}[h]
\centering
\includegraphics[width=0.95\linewidth]{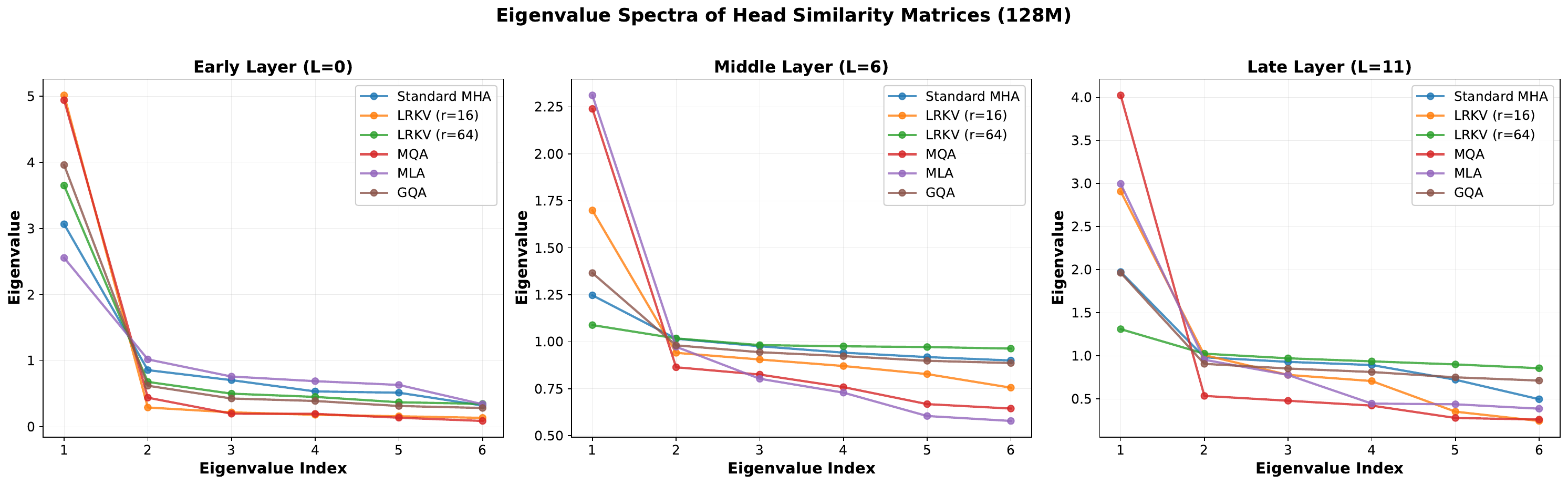}
\caption{
\textbf{Uncentered eigenvalue spectra at 128M scale.} With 6 heads, the eigenvalue structure is more pronounced. LRKV (r=64) tracks Standard MHA closely, while LRKV (r=16) shows slightly elevated leading eigenvalues in later layers, consistent with reduced effective rank. MQA exhibits the most concentrated spectra in the uncentered view, with a dominant leading eigenvalue suggesting strong head redundancy—but this is largely due to the shared KV mean direction rather than true loss of diversity, as PCA analysis reveals.
}
\label{fig:eigenvalue_spectra_128M_uncentered}
\end{figure}

\subsection{Head Similarity Heatmaps}

\begin{figure}[h]
\centering
\includegraphics[width=0.95\linewidth]{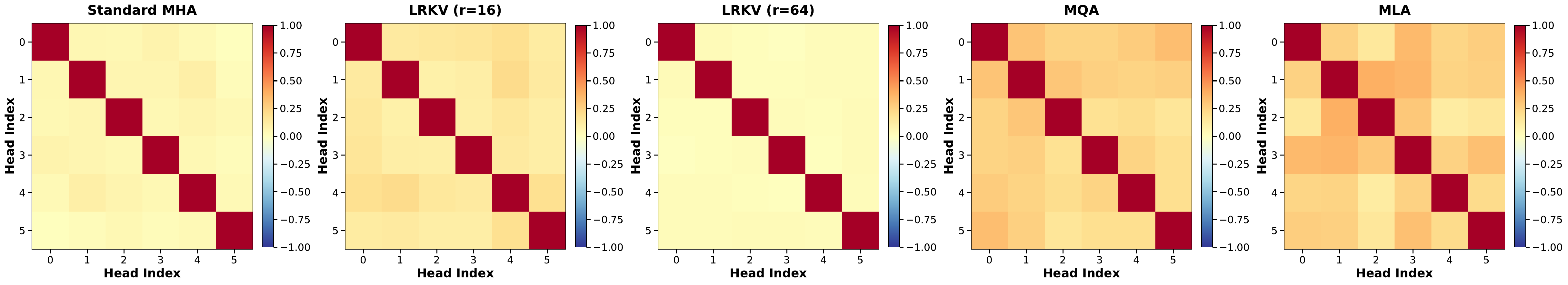}
\caption{
\textbf{Head similarity matrices at 128M scale.} With fewer heads (6 vs 18), individual head relationships are more visible. LRKV (r=64) shows similarity structure nearly identical to Standard MHA, with moderate positive correlations but no dominant clusters. LRKV (r=16) exhibits slightly stronger correlations, consistent with its reduced effective rank. The mean off-diagonal similarity for LRKV (r=64) is within 0.02--0.04 of Standard MHA across layers, providing quantitative confirmation of the visual similarity. These uncentered heatmaps complement the PCA-based analysis by showing the raw pairwise similarities before mean removal.
}
\label{fig:similarity_heatmaps_128M}
\end{figure}

The heatmaps provide direct visualization of head relationships in the uncentered similarity space, complementing the PCA-based aggregate metrics. Key observations:

\begin{itemize}
\item \textbf{LRKV preserves correlation structure:} At both scales, LRKV's similarity matrices closely match Standard MHA's patterns. Off-diagonal entries (head-to-head similarities) show comparable magnitudes and spatial distribution, indicating that low-rank factorization does not artificially increase or decrease head correlations in the uncentered space.

\item \textbf{No emergent clustering:} Standard MHA shows diffuse correlations without strong block structure (aside from diagonal dominance). LRKV maintains this property, suggesting heads remain independently specialized rather than forming redundant groups, even before centering removes shared mean structure.

\item \textbf{GQA shows group structure:} At 2.5B scale with 6 KV groups of 3 heads each, GQA's heatmap reveals visible within-group structure—heads sharing KV projections exhibit stronger mutual similarity. This validates that partial KV sharing induces measurable correlation visible in raw similarity space.

\item \textbf{Quantitative confirmation:} The mean off-diagonal similarity for LRKV (r=64) is within 0.02--0.04 of Standard MHA across layers, confirming the visual similarity is not merely qualitative. This tight correspondence holds in both uncentered and PCA-based analysis.
\end{itemize}

\subsection{Key Findings from PCA Analysis}

\paragraph{1. LRKV preserves head independence within 1\% of standard attention.}
At 2.5B scale, LRKV achieves 93.5\% PCA-based effective rank versus 94.0\% for Standard MHA (0.5pp gap). This near-perfect preservation occurs despite 52.6\% KV cache size, validating that rank-64 residuals provide sufficient capacity for head specialization. The consistency across all 36 layers demonstrates depth-invariant factorization quality.

\paragraph{2. The MQA compensation effect.}
MQA's 4.8pp improvement from uncentered (86.2\%) to PCA-based (91.0\%) effective rank at 2.5B scale reveals a previously unknown mechanism: forced KV sharing creates a strong mean bilinear form, but heads compensate by diversifying query projections around this mean. The centered analysis isolates this true independence, showing MQA is less constrained than prior metrics suggest. This effect is consistent at 128M scale (5.8pp improvement), confirming it is a fundamental property of MQA rather than a scale-specific artifact.

\paragraph{3. MLA sits between GQA and MQA in diversity space.}
MLA achieves 91.6\% PCA-based effective rank at 2.5B, placing it between GQA (91.7\%) and MQA (91.0\%). This makes architectural sense: MLA's latent bottleneck constrains heads more than GQA's group sharing but less than MQA's complete sharing, with per-head decompression allowing specialization after the bottleneck. At 128M scale, MLA shows 78.5\% effective rank, slightly below MQA (78.4\%), suggesting the latent bottleneck is more constraining at smaller scales where bottleneck capacity is limited.

\paragraph{4. Scale-dependent behavior.}
The gap between uncentered and PCA-based metrics shrinks from 128M to 2.5B (e.g., LRKV: 13.2pp gap at 128M vs 4.8pp at 2.5B), suggesting larger models develop stronger head specialization around shared structure rather than purely independent representations. This transition indicates that model scale affects not just capacity but the fundamental organization principle of attention heads.

\paragraph{5. Rank selection determines capacity.}
Comparing LRKV configurations reveals that appropriate rank is critical. At 128M scale, r=16 (12.5\% of head dimension) achieves 81.5\% PCA-based effective rank, while r=64 (50\%) achieves 83.0\%—a modest 1.5pp improvement in diversity but substantial performance difference (0.881 vs 0.875 BPB). This confirms that representational capacity, not geometric properties, determines quality: even moderate diversity can yield strong performance if the rank provides sufficient degrees of freedom.

\subsection{Comparison to Prior Diversity Metrics}

Our PCA-based approach differs from prior work in several key respects:

\paragraph{CKA/SVCCA \citep{kornblith2019similarity,raghu2017svcca}.} These methods compare activation similarities (representations), not functional operators. While CKA uses centered Gram matrices (providing the "C" in "Centered Kernel Alignment"), it analyzes activation space rather than weight-space bilinear forms. Our contribution is applying the centering principle to the space of attention operators themselves, revealing compensation mechanisms invisible when analyzing activations.

\paragraph{Attention pattern similarity \citep{michel2019sixteen,voita2019analyzing}.} These behavioral metrics measure which tokens heads attend to on specific inputs. They depend on input data distribution and don't directly measure the representational capacity encoded in the parameterization. Our approach analyzes the parameterization itself, providing a data-independent measure of potential diversity.

\paragraph{Raw weight comparison.} Directly comparing $\mathbf{W}_h^K$ or $\mathbf{W}_h^Q$ matrices is not gauge-invariant: arbitrary per-head rotations that preserve attention function can make identical heads appear different or vice versa. The bilinear form $\mathbf{W}^Q (\mathbf{W}^K)^\top$ is the minimal gauge-invariant object for comparison.

\paragraph{Uncentered Gram matrices.}
Several prior analyses\citep{zhang2024improving} form head similarity matrices (often using uncentered correlations or kernel similarities) and interpret their spectra directly; our contribution is to apply explicit kernel centering in the space of gauge-invariant attention operators, which separates shared mean structure from variance around the mean.
This conflates mean structure (shared baselines) with spread (true diversity), masking compensation effects like MQA's query specialization. Our centered analysis isolates intrinsic dimensionality independent of mean direction.

The combination of (i) gauge-invariant bilinear forms, (ii) centered Gram matrix (kernel PCA), and (iii) PCA interpretation as variance explained provides, to our knowledge, the first principled operator-space analysis of transformer head diversity. This methodology reveals phenomena invisible to prior approaches, such as the MQA compensation effect and the precise quantification of LRKV's near-perfect diversity preservation.

\subsection{Methodological Limitations and Extensions}

\paragraph{Limitations.}
Our analysis focuses on final trained checkpoints and does not track how the factorization evolves during training. The PCA-based metric measures potential diversity encoded in the parameterization but not behavioral diversity on specific inputs. The gauge-invariant similarity metric is one of many possible inner products on bilinear forms; alternative metrics might reveal additional structure.

\paragraph{Future extensions.}
Several directions merit exploration: (1) Analyzing PCA eigenvalue dynamics during pretraining to understand how shared and head-specific structure co-evolve. (2) Comparing parameterization-based diversity (this work) with input-dependent behavioral diversity to understand their relationship. (3) Extending the framework to cross-attention in encoder-decoder models, where query and key-value heads may have different sizes. (4) Investigating whether PCA-guided initialization or regularization can improve training dynamics by explicitly encouraging high-variance factorizations.

\section{Experimental Setup (Full Details)}
\label{appendix:experimental_setup}

\begin{table}[h]
\centering
\caption{\textbf{Mathematical comparison of attention mechanisms.}
We compare how different mechanisms parameterize key/value projections and where
compression is applied. Here $X\in\mathbb{R}^{T\times d}$ is the token sequence,
$d_h=d/H$ is the head dimension, $H$ is the number of heads, and $r,d_c\ll d_h$.}
\label{tab:attention_math_comparison}
\resizebox{\textwidth}{!}{%
\small
\begin{tabular}{l p{6cm} p{4.5cm} p{4.5cm}}
\toprule
\textbf{Method} & \textbf{KV projection form} & \textbf{Rank constraint} & \textbf{Compression axis} \\
\midrule
\textbf{Standard MHA}
&
$W_h^{K,V} \in \mathbb{R}^{d\times d_h}$ (independent per head)
&
None (full rank)
&
None \\
\midrule
\textbf{MQA}
&
$W_h^{K,V} = W_{\mathrm{shared}}^{K,V} \;\;\forall h$
&
$\mathrm{rank}(W_h^{K,V}) = \mathrm{rank}(W_{\mathrm{shared}})$
&
Across heads (complete sharing) \\
\midrule
\textbf{GQA}
&
$W_h^{K,V} = W_{g(h)}^{K,V}$ for group $g(h)$
&
Full rank within group
&
Across heads (group-wise sharing) \\
\midrule
\textbf{MLA}
&
$W_h^{K,V} = W_{\mathrm{down}} W_{\mathrm{up},h}^{K,V}$
&
$\mathrm{rank}(W_h^{K,V}) \le d_c$
&
Across tokens (latent bottleneck) \\
\midrule
\textbf{LRKV (ours)}
&
$W_h^{K,V} = W_{\mathrm{shared}}^{K,V} + U_h^{K,V} (B_h^{K,V})^\top$
&
$\mathrm{rank}(W_h^{K,V} - W_{\mathrm{shared}}^{K,V}) \le r$
&
Across heads (additive low-rank deviations) \\
\bottomrule
\end{tabular}%
}
\end{table}

\subsection{Pretraining Configuration}

We pretrain from scratch a family of decoder-only Transformer language models with parameter counts ranging from \textbf{128M to 6.3B} on the \textbf{FineWeb-Edu} dataset~\citep{fineweb2024}.
Unless otherwise stated, model scale (e.g., 128M, 2.5B, 6.3B) refers to the parameter count of the \emph{base architecture with standard multi-head attention (MHA)}. Alternative attention mechanisms (LRKV, MQA, GQA, MLA) replace only the attention module while keeping all other architectural components fixed, which will result in reduced total parameter count due to reparameterized key-value projections.

\paragraph{Model Architecture Summary.}
Table~\ref{tab:model_architectures} provides complete architectural specifications for all model scales evaluated in this work.

\begin{table}[h]
\centering
\caption{\textbf{Complete model architectural specifications.} All models use $d_{\text{head}}=128$ and FFN expansion ratio of 4 (i.e., $d_{\text{FFN}} = 4 \times d_{\text{model}}$).}
\label{tab:model_architectures}
\small
\begin{tabular}{lccccc}
\toprule
\textbf{Scale} & \textbf{Layers} & \textbf{Heads} & \textbf{$d_{\text{model}}$} & \textbf{$d_{\text{FFN}}$} & \textbf{Vocab Size} \\
\midrule
128M & 12 & 6 & 768 & 3072 & 50304 \\
512M & 24 & 12 & 1536 & 6144 & 50304 \\
1.2B & 24 & 12 & 1536 & 6144 & 50304 \\
2.5B & 18 & 18 & 2304 & 9216 & 50304 \\
6.3B & 32 & 32 & 4096 & 16384 & 50304 \\
\bottomrule
\end{tabular}
\end{table}

\paragraph{Training Configuration Summary.}
Table~\ref{tab:training_config} provides complete training configuration details across all model scales.

\begin{table}[h]
\centering
\caption{\textbf{Training configuration across model scales.} Batch size refers to number of sequences; total tokens per batch = batch size $\times$ context length. All models trained on FineWeb-Edu with Muon+AdamW optimization.}
\label{tab:training_config}
\small
\begin{tabular}{lcccccc}
\toprule
\textbf{Scale} & \textbf{Context} & \textbf{Batch Size} & \textbf{Tokens/Batch} & \textbf{Total Tokens} & \textbf{Compute (EF)} & \textbf{Purpose} \\
\midrule
128M & 2048 & 16 & 32,768 & 100B & 31.0 & Ablations \\
512M & 8192 & 4 & 32,768 & 50B & 153.6 & Long context \\
1.2B & 2048 & 8 (GA=2) & 32,768 & 50B & 304.7 & Scaling study \\
2.5B & 2048 & 8 (GA=2) & 32,768 & 50B & 635.2 & Main experiments \\
6.3B & 2048 & 8 (GA=2) & 32,768 & 50B & 1603.3 & Scaling study \\
\bottomrule
\end{tabular}
\caption*{\small GA=gradient accumulation steps. EF=ExaFLOPs. 128M overtrained (100B tokens) for low-variance ablations; others use compute-near-optimal budgets~\citep{hoffmann2022training}.}
\end{table}

\paragraph{Optimization Details.}
We use a hybrid optimizer strategy: the \textbf{Muon optimizer}~\citep{muon2024} for Transformer weight matrices and \textbf{AdamW} for embeddings, with component-specific learning rates (matrix: 0.02, embedding: 0.2, unembedding: 0.004). All models are optimized using cosine learning rate decay with linear warmup (2000 steps), weight decay 0.1, and gradient clipping at 1.0.

\paragraph{Hardware and Precision.}
All experiments are run on a single \textbf{8$\times$H200 GPU} node in mixed-precision (bfloat16) mode. Full architectural and optimization hyperparameters, including learning rate schedules and detailed optimizer settings, are provided in Appendix~\ref{appendix:hyperparams}.

\paragraph{Evaluation Metrics.}
Training progress is monitored using the \textbf{cross-entropy loss} and \textbf{bits-per-byte (BPB)} on held-out validation split of FineWeb-Edu. These metrics allow consistent comparison across model scales and serve as the primary indicators of data efficiency and convergence quality.

\subsection{Mid-training Configuration}

After pretraining, we perform \textbf{supervised fine-tuning} (mid-training) on a curated mixture of instructional data to improve downstream task performance.
The mid-training dataset consists of three components:
\begin{enumerate}
\item \textbf{SmolTalk}~\citep{allal2024smoltalk} with 460K conversational examples for general instruction-following
\item \textbf{MMLU auxiliary train}~\citep{hendrycks2021mmlu} with 100K multiple-choice questions spanning diverse academic subjects
\item \textbf{GSM8K}~\citep{cobbe2021gsm8k} with 8K grade-school math problems including calculator tool use
\end{enumerate}

This yields a total training mixture of approximately \textbf{568K examples}.
Validation is performed on held-out test splits using proportional sampling (24K SmolTalk, 5.2K MMLU, 420 GSM8K examples).

We train for \textbf{one full epoch} over the mid-training mixture with a batch size of \textbf{524,288 tokens} and sequence length of 2048. We use a decayed linear schedule for the learning rate, using AdamW for the embeddings and Muon for the remaining parameters.
Models are evaluated every 150 steps using bits-per-byte on the validation set.
All mid-training runs use the same hyperparameters across different attention mechanisms to ensure fair comparison.

\subsection{Attention Mechanism Configurations}

To evaluate the effectiveness of low-rank KV factorization, we compare LRKV against four baseline attention mechanisms across all model scales. Table~\ref{tab:attention_configs} provides complete configuration details for all methods and scales.

\begin{table}[h]
\centering
\caption{\textbf{Attention mechanism configurations across model scales.} KV cache percentages are relative to Standard MHA. All models use $d_{\text{head}}=128$ and 2048-token context (except 512M long-context with 8192 tokens).}
\label{tab:attention_configs}
\resizebox{\textwidth}{!}{%
\small
\begin{tabular}{l|ccc|ccc|ccc}
\toprule
& \multicolumn{3}{c|}{\textbf{Standard}} & \multicolumn{3}{c|}{\textbf{MQA / GQA}} & \multicolumn{3}{c}{\textbf{MLA / LRKV}} \\
\cmidrule(lr){2-4} \cmidrule(lr){5-7} \cmidrule(lr){8-10}
\textbf{Scale} & \textbf{Heads} & \textbf{KV Heads} & \textbf{Cache} & \textbf{MQA KV} & \textbf{GQA KV} & \textbf{GQA Cache} & \textbf{MLA $d_c$} & \textbf{LRKV $r$} & \textbf{LRKV Cache} \\
\midrule
128M & 6 & 6 & 100\% & 1 (16.7\%) & 3 (50\%) & 50\% & 128 (8.3\%) & 46 & 52.6\% \\
512M & 12 & 12 & 100\% & 1 (8.3\%) & 4 (33.3\%) & 33.3\% & 256 (8.3\%) & 51 & 48.2\% \\
1.2B & 12 & 12 & 100\% & 1 (8.3\%) & 4 (33.3\%) & 33.3\% & 256 (8.3\%) & 51 & 48.2\% \\
2.5B & 18 & 18 & 100\% & 1 (5.6\%) & 6 (33.3\%) & 33.3\% & 384 (8.3\%) & 55 & 48.5\% \\
6.3B & 32 & 32 & 100\% & 1 (3.1\%) & 2 (6.3\%) & 6.3\% & 1024 (12.5\%) & 54 & 45.3\% \\
\midrule
512M-8K$^\dagger$ & 12 & 12 & 100\% & 1 (8.3\%) & 4 (33.3\%) & 33.3\% & 256 (8.3\%) & 51 & 48.2\% \\
\bottomrule
\end{tabular}%
}
\caption*{\small $^\dagger$512M-8K denotes the long-context configuration with 8192-token sequences. All other models use 2048-token context.}
\end{table}

\paragraph{KV Cache Calculation.}
All KV cache percentages are reported relative to Standard MHA's memory usage. For LRKV, the reported cache size assumes an optimized implementation that caches the shared projection $W_{\text{shared}}$ and per-head low-rank latents $R_h^{K,V}$ rather than the fully reconstructed key-value matrices, following the theoretical analysis in Section~\ref{sec:methodology}. The cache ratio follows: $\frac{d_h + Hr}{Hd_h} = \frac{1}{H} + \frac{r}{d_h}$.

This configuration balances memory efficiency with model expressiveness: LRKV uses approximately half the KV cache of Standard MHA while maintaining full representational capacity through low-rank adaptation, outperforming both memory-minimal approaches (MQA, MLA) and full attention in downstream task performance.

\subsection{Rank Ablation Experiments}

To systematically evaluate how residual rank affects LRKV's performance-memory tradeoff, we conduct comprehensive ablation studies at the 128M scale. These experiments isolate the effect of rank selection while controlling for all other architectural and training factors.

\paragraph{LRKV Rank Ablation.}
We train five 128M parameter models with LRKV using ranks $r \in \{8, 16, 32, 64, 128\}$ on 100B tokens of FineWeb-Edu. All models use identical architecture (6 heads, $d_{\text{model}}=768$, $d_{\text{head}}=128$), optimization hyperparameters, and training schedule. Table~\ref{tab:rank_ablation_results} presents complete results.

\begin{table}[h]
\centering
\caption{\textbf{LRKV rank ablation and MLA latent dimension comparison (128M scale, 100B tokens).} All models use 6 heads with $d_h=128$. Cache percentages relative to Standard MHA baseline.}
\label{tab:rank_ablation_results}
\small
\begin{tabular}{l|ccccc}
\toprule
\textbf{Configuration} & \textbf{Rank/Latent} & \textbf{KV Cache} & \textbf{CE Loss} $\downarrow$ & \textbf{BPB} $\downarrow$ & \textbf{Notes} \\
\midrule
\multicolumn{6}{c}{\textit{LRKV Rank Ablation}} \\
\midrule
LRKV-8 & $r=8$ & 22.9\% & 2.908 & 0.881 & Insufficient capacity \\
LRKV-16 & $r=16$ & 29.2\% & 2.905 & 0.880 & Below MHA \\
LRKV-32 & $r=32$ & 41.7\% & 2.896 & 0.877 & Approaching MHA \\
LRKV-64 & $r=64$ & 66.7\% & 2.888 & 0.875 & Exceeds MHA \\
LRKV-128 & $r=128$ & 116.7\% & 2.877 & 0.873 & Best, but $>$100\% cache \\
\midrule
\multicolumn{6}{c}{\textit{MLA Latent Dimension Ablation}} \\
\midrule
MLA-128 & $d_c=128$ & 8.3\% & 2.901 & 0.876 & Minimal memory \\
MLA-192 & $d_c=192$ & 12-5\% & 2.898 & 0.876 & -- \\
MLA-384 & $d_c=384$ & 25.0\% & 2.894 & 0.875 & Comparable to LRKV-64 \\
\midrule
\multicolumn{6}{c}{\textit{Baselines}} \\
\midrule
Standard MHA & 6 KV heads & 100\% & 2.903 & 0.878 & Baseline \\
GQA & 3 groups & 50.0\% & 2.918 & 0.883 & -- \\
MQA & 1 KV head & 16.7\% & 2.929 & 0.886 & Severe quality loss \\
\bottomrule
\end{tabular}
\end{table}

\paragraph{Key Findings.}
The ablation reveals that performance improves monotonically with rank, confirming that representational capacity (not geometric properties of the factorization) determines modeling quality. Ranks around $r \approx 0.36$--$0.43 \times d_h$ (46--55 for $d_h=128$) provide the optimal tradeoff: sufficient capacity to exceed Standard MHA performance while achieving substantial cache reduction.

Comparing methods at matched cache budgets reveals LRKV's architectural advantage: LRKV at 41.7\% cache ($r=32$, BPB=0.877) slightly underperforms MLA at 50.0\% cache ($d_c=384$, BPB=0.875), but LRKV at 66.7\% cache ($r=64$, BPB=0.875) matches MLA-384 performance. However, LRKV avoids the sequence-length-dependent reconstruction overhead inherent to MLA's latent compression (see~\autoref{appendix:hyperparams}), providing better inference latency characteristics at long context lengths.

All ablation models are trained for 100B tokens (rather than compute-optimal allocation) to reduce variance and enable high-confidence comparison of architectural capacity at fixed training budget.

\subsection{Long Context Experiments}

To evaluate LRKV's effectiveness at extended sequence lengths, we conduct 8192-token context experiments using 512M parameter models trained on 50B tokens (153.6 ExaFLOPs).

\paragraph{Configuration.}
The 512M models use 12 attention heads with $d_{\text{model}}=1536$ and $d_{\text{head}}=128$, with 24 layers matching the 1.2B scale architecture. We compare LRKV ($r=51$, 48.2\% cache), GQA (4 groups, 33.3\% cache), MLA ($d_c=256$, 8.3\% cache), and MQA (8.3\% cache) against Standard MHA baseline.

\paragraph{Batching Strategy.}
To maintain computational throughput while accommodating 4$\times$ longer contexts, we adjust batch size to keep total tokens per batch constant. Standard 2048-context runs use batch size 16 (32,768 tokens per batch); 8192-context runs use batch size 4 (32,768 tokens per batch). This ensures comparable gradient noise and optimizer dynamics while scaling to long contexts.

\paragraph{Results.}
Table~\ref{tab:long_context_results} presents validation performance at 8192-token context length after 50B tokens (153.6 ExaFLOPs).

\begin{table}[h]
\centering
\caption{\textbf{Long context performance (512M models, 8K context, 50B tokens).} Degradation computed relative to MHA baseline.}
\label{tab:long_context_results}
\small
\begin{tabular}{lcccc}
\toprule
\textbf{Method} & \textbf{KV Cache} & \textbf{CE Loss} $\downarrow$ & \textbf{BPB} $\downarrow$ & \textbf{vs. MHA} \\
\midrule
Standard MHA & 100\% & 2.740 & 0.494 & baseline \\
\textbf{LRKV} & 48.2\% & \textbf{2.665} & \textbf{0.481} & \textbf{-2.7\%} \\
GQA & 33.3\% & 2.704 & 0.488 & -1.3\% \\
MLA & 8.3\% & 2.714 & 0.489 & -0.9\% \\
MQA & 8.3\% & 2.728 & 0.492 & -0.4\% \\
\bottomrule
\end{tabular}
\end{table}

LRKV achieves the strongest performance, outperforming Standard MHA by 2.7\% while using less than half the KV cache (48.2\% vs 100\%). All KV-efficient methods (LRKV, GQA, MLA, MQA) outperform the full Standard MHA baseline at 8K context, suggesting that some degree of KV compression may provide implicit regularization benefits at longer sequences. However, LRKV's 2.7\% advantage over MHA (compared to GQA's 1.3\%, MLA's 0.9\%, and MQA's 0.4\%) demonstrates that head-specific low-rank residuals are particularly effective for maintaining quality at extended context lengths.

These results validate LRKV's applicability to long-context scenarios where memory efficiency is critical, demonstrating that low-rank factorization preserves head diversity more effectively than aggressive sharing (MQA, GQA) or latent compression (MLA) approaches at extended sequence lengths.

\subsection{Downstream Evaluation Tasks}

After midtraining, we evaluate all models on five standard benchmarks:
\begin{itemize}
\item \textbf{ARC-Easy and ARC-Challenge}~\citep{clark2018arc}: Question answering requiring scientific reasoning
\item \textbf{MMLU}~\citep{hendrycks2021mmlu}: Multi-task language understanding across 57 subjects
\item \textbf{GSM8K}~\citep{cobbe2021gsm8k}: Grade-school math word problems
\item \textbf{HumanEval}~\citep{chen2021humaneval}: Python code generation from docstrings
\end{itemize}

All evaluations follow standard zero-shot or few-shot protocols as specified in the original benchmark papers.

\section{Pretraining-Downstream Correlation Analysis}
\label{appendix:pretraining_downstream_correlation}

\begin{figure}[h]
\centering
\includegraphics[width=0.6\linewidth]{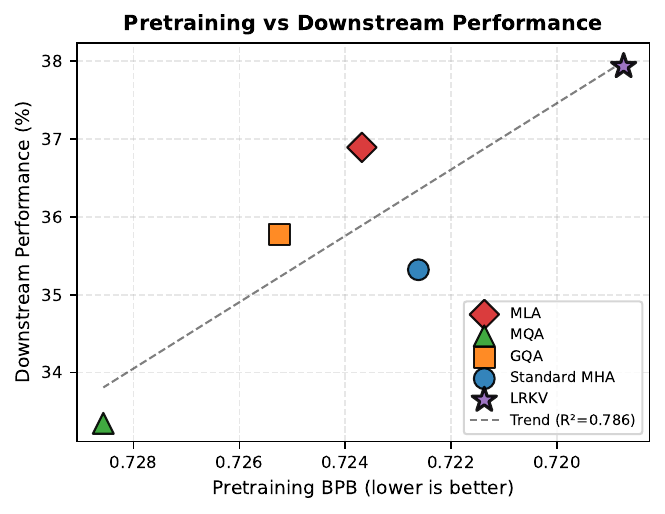}
\caption{\textbf{Pretraining quality predicts downstream performance.} Scatter plot relating pretraining validation BPB (x-axis, inverted) to downstream combined performance (y-axis) across all attention mechanisms at 2.5B scale. A strong positive correlation emerges ($R²=0.786$), demonstrating that models with better pretraining performance consistently achieve higher downstream scores. LRKV occupies the optimal position with both the lowest pretraining BPB (0.719) and highest downstream performance (37.9\%), while Standard MHA achieves 0.723 BPB and 35.3\% downstream.}
\label{fig:pretraining_downstream}
\end{figure}

To validate the relationship between pretraining quality and downstream performance, we analyze the correlation between validation BPB and combined benchmark scores across all architectures after midtraining (Figure~\ref{fig:pretraining_downstream}). The strong positive correlation (R²=0.786) demonstrates that models achieving lower pretraining loss consistently deliver superior downstream performance. LRKV achieves the lowest pretraining BPB (0.719) and highest downstream score (37.9\%), while Standard MHA attains 0.723 BPB and 35.3\% downstream.

This correlation suggests that architectural improvements that enhance language modeling performance are not orthogonal to, but rather aligned with, practical task capabilities. The consistent 2.6 percentage point downstream advantage LRKV maintains over Standard MHA directly stems from its BPB pretraining improvement, confirming that low-rank KV factorization provides fundamental capacity gains that manifest across both pretraining and downstream evaluation regimes.

This relationship validates that LRKV's architectural improvements during pretraining directly translate to enhanced downstream capabilities, confirming the effectiveness of low-rank KV factorization for both language modeling and task-specific adaptation.

\section{Additional Experimental Details and Computational Analysis}\label{appendix:hyperparams}

\subsection{Batching and Optimization Hyperparameters}

For models up to 256M parameters, we use a global batch size of \textbf{16} sequences (2048 tokens each).
For the larger 1B--6.3B models, we reduce the physical batch size to \textbf{8} sequences and apply \textbf{gradient accumulation of 2 steps} to maintain an equivalent effective batch size.
This setup ensures comparable gradient noise scale and optimizer dynamics across all model sizes. All runs employ weight decay of 0.1, Adam $\beta_1 = 0.9$, $\beta_2 = 0.95$, and gradient clipping at 1.0.

\subsection{Memory-Computation Trade-offs in Attention Mechanisms}

Modern attention mechanisms present a fundamental trade-off between cache memory (storage) and per-token computation (latency). We analyze this trade-off across standard attention, query-sharing variants, and compression-based approaches.

\paragraph{The Compression Paradigm: Lossy vs. Lossless.}
Attention mechanisms can be understood through the lens of compression theory:

\begin{itemize}
\item \textbf{Multi-Latent Attention (MLA)} implements \emph{lossy compression}: it compresses K/V representations into a low-dimensional latent space $Z = XW_{\text{down}}$, caching only $Z$. During generation, full K/V need not be explicitly materialized; instead, interaction with the cached latent requires per-step projection overhead, e.g.,
$qK^\top = (qW_{\text{up}}^K)Z^\top$ and $aV = (aZ)W_{\text{up}}^V$.
This introduces additional per-token computation proportional to the latent dimension, and scales linearly with sequence length through dot products with $Z^\top$.

\item \textbf{LRKV} implements \emph{near-lossless compression with additive structure}: it caches both shared full-rank features ($\bar{K}, \bar{V}$) and compact per-head latents ($R_h^K, R_h^V$). It avoids explicitly materializing per-head $K_h,V_h$ tensors; instead, attention can be computed using associative forms (Eq.~\ref{eq:lrkv_logits}--\ref{eq:lrkv_values}). Beyond the unavoidable $O(T)$ scan over cached tokens in attention, LRKV introduces only $O(H r d_h)$ additional per-step projection work that does not grow with $T$.

\end{itemize}

This distinction is critical for autoregressive generation: both MLA and LRKV incur the unavoidable $O(T)$ cost to compare against cached tokens, but MLA introduces additional per-step projection work tied to the latent bottleneck, whereas LRKV's extra work depends only on $r$ and does not increase with $T$.

\paragraph{Quantitative Analysis: Memory vs.\ Latency.}
We compare the memory footprint and inference-time computational characteristics of
different attention mechanisms during autoregressive generation.
Table~\ref{tab:memory_latency_tradeoff} reports (1) the KV cache size stored per token and (2) the \emph{additional reconstruction overhead} required per generation step, \emph{beyond} the standard $O(T)$ attention computation inherent to cached autoregressive decoding.

\begin{table}[h]
\centering
\caption{
Memory and inference-time characteristics of attention mechanisms.
Cache size is measured per token.
Reconstruction overhead denotes \emph{additional} computation beyond standard
$O(T)$ cached attention.
}
\label{tab:memory_latency_tradeoff}
\small
\begin{tabular}{lccc}
\toprule
\textbf{Method} &
\textbf{Cache / Token} &
\textbf{Extra Cost / Step} &
\textbf{$T$-Dependence} \\
\midrule
Standard Attn. & $2Hd_h$ & $O(1)$ & None \\
MQA & $2d_h$ & $O(1)$ & None \\
GQA ($G{=}3$) & $2Gd_h$ & $O(1)$ & None \\
MLA ($d_c{=}128$) & $d_c$ & $O(T\,d_c d_h)$ & Linear \\
\textbf{LRKV} ($r{=}16$) & $2(d_h{+}Hr)$ & $O(Hr d_h)$ & None \\
\bottomrule
\end{tabular}

\vspace{0.4em}
\caption*{\small
Extra cost refers to reconstruction needed to obtain per-head K/V representations.
MLA requires latent-to-head expansion over the cached sequence unless expanded K/V
are stored or fused kernels are used.
}
\end{table}

\textbf{Key observations.}
\begin{enumerate}
  \item Standard attention, MQA, and GQA incur no additional reconstruction cost at
  inference time, as per-head or shared key/value tensors are directly cached.
  \item MLA achieves the smallest cache footprint but introduces an additional
  latent-to-head expansion step whose cost scales with the cached sequence length $T$, unless expanded representations are stored (which increases memory) or specialized fused kernels are used.

  \item LRKV trades a modest increase in cache size relative to MLA for
  sequence-length-independent reconstruction overhead. This avoids $T$-dependent
  expansion during generation while preserving per-head expressivity through low-rank residuals.

  \item As a result, LRKV occupies an intermediate point in the memory--latency design space: it substantially reduces KV memory relative to full attention while avoiding the sequence-dependent reconstruction overhead introduced by more aggressive compression schemes.
\end{enumerate}

\section{KV Cache Memory Measurements}
\label{appendix:memory_profiling}

This appendix presents detailed measurements of KV cache memory usage during inference for all attention mechanisms evaluated in this work.

\subsection{Measurement Methodology}

All memory measurements are obtained by instantiating cache tensors with the exact shapes used during inference and measuring their memory footprint using PyTorch's \texttt{tensor.numel()} and \texttt{element\_size()} methods. Measurements use bfloat16 precision (2 bytes per element) with batch size 1 and sequence length 2048 tokens, matching our experimental configuration.

\subsection{Cache Structure by Method}

Table~\ref{tab:cache_structure} summarizes the cache structure for each attention mechanism, showing tensor shapes and memory formulas for inference with batch size $B$, sequence length $T$, number of layers $L$, number of heads $H$, head dimension $d_h$, group size $G$, latent dimension $d_c$, and rank $r$.

\begin{table}[h]
\centering
\caption{\textbf{KV cache structure by attention mechanism.} All measurements use bfloat16 precision (2 bytes per element).}
\label{tab:cache_structure}
\small
\begin{tabular}{l p{5cm} p{5.5cm}}
\toprule
\textbf{Method} & \textbf{Cached Tensor/s} & \textbf{Total Memory (bytes)} \\
\midrule
Standard MHA
& Shape: $(L, B, H, T, d_h)$
& $2 \times L \times B \times H \times T \times d_h \times 2$ \\
\midrule
MQA
& Shape: $(L, B, 1, T, d_h)$
& $2 \times L \times B \times T \times d_h \times 2$ \\
\midrule
GQA
& Shape: $(L, B, H/G, T, d_h)$
& $2 \times L \times B \times (H/G) \times T \times d_h \times 2$ \\
\midrule
MLA
& Shape: $(L, B, T, d_c)$ (latents)
& $L \times B \times T \times d_c \times 2$ \\
\midrule
LRKV (Optimized)
& Shared: $(L, B, T, d_h)$\newline Latents: $(L, B, H, T, r)$
& $2 \times L \times B \times T \times (d_h + H \times r) \times 2$ \\
\bottomrule
\end{tabular}
\end{table}

\subsection{Measured Results}

Table~\ref{tab:measured_kv_cache} presents measured memory usage across all three model scales. The optimized LRKV implementation (using \texttt{LRKVCache} from \texttt{lrkv\_attention\_fused.py}) achieves substantial memory savings compared to standard MHA while maintaining full representational capacity.

\begin{table}[h]
\centering
\caption{\textbf{Measured KV cache memory during inference.} All values measured for T=2048 tokens, batch size 1, bfloat16 precision, using rank $r=64$ for LRKV and latent dimensions as specified in main paper.}
\label{tab:measured_kv_cache}
\small
\begin{tabular}{l|ccc}
\toprule
\textbf{Method} & \textbf{128M} & \textbf{2.5B} & \textbf{6.3B} \\
\midrule
Standard MHA & 72.0 MB & 648.0 MB & 1152.0 MB \\
MQA & 12.0 MB (16.7\%) & 36.0 MB (5.6\%) & 36.0 MB (3.1\%) \\
GQA & 36.0 MB (50.0\%) & 216.0 MB (33.3\%) & 72.0 MB (6.2\%) \\
MLA & 24.0 MB (33.3\%) & 216.0 MB (33.3\%) & 288.0 MB (25.0\%) \\
\textbf{LRKV} & \textbf{48.0 MB (52.6\%)} & \textbf{360.0 MB (48.4\%)} & \textbf{612.0 MB (45.1\%)} \\
\bottomrule
\end{tabular}
\end{table}

\subsection{Scale-Dependent Efficiency}

LRKV's cache percentage relative to standard MHA improves with model scale:
\begin{align*}
\text{Ratio} &= \frac{d_h + H \cdot r}{H \cdot d_h} = \frac{1}{H} + \frac{r}{d_h} \\
\text{128M (H=6):} &\quad \frac{1}{6} + \frac{64}{128} = 0.526 = 52.6\% \\
\text{2.5B (H=18):} &\quad \frac{1}{18} + \frac{64}{128} = 0.484 = 48.4\% \\
\text{6.3B (H=32):} &\quad \frac{1}{32} + \frac{64}{128} = 0.451 = 45.1\%
\end{align*}

This demonstrates that LRKV's efficiency \emph{increases} with model scale: larger models with more heads benefit more from the shared component, achieving greater relative memory savings while maintaining fixed rank $r=64$.

\section{Limitations and Future Work}
\label{appendix:limitations_future}

\subsection{Limitations}

Our experiments focus on decoder-only language models at scales up to 6.3B parameters. The design space for large-scale Transformer training is extremely large, spanning model size, training data, optimization schedules, attention mechanisms, and hardware/software configurations, many of which interact in nontrivial ways. As a result, it is infeasible to exhaustively benchmark all combinations or to claim that any single configuration is universally optimal across setups.

Our empirical evaluation therefore samples this space using widely adopted architectures, training recipes, and system implementations on modern accelerator hardware. While this provides a representative and practically relevant assessment, different choices of model scale, data regime, training duration, or system optimizations may shift the precise efficiency--performance trade-offs.

To facilitate independent validation and extension, we will release our training code and configuration details, enabling replication and exploration across alternative setups. We also note that the relative efficiency of different KV compression strategies may vary with hardware characteristics such as memory bandwidth, kernel fusion, and cache behavior.

\end{document}